\documentclass[10pt,twocolumn,letterpaper]{article}

\usepackage[pagenumbers]{wacv} 

\usepackage{soul}
\usepackage{multirow}

\usepackage{array}

\usepackage{graphicx}
\usepackage{amsmath}
\usepackage{amssymb}
\usepackage{booktabs}
\usepackage[symbol]{footmisc}

\definecolor{wacvblue}{rgb}{0.21,0.49,0.74}
\usepackage[pagebackref,breaklinks,colorlinks,allcolors=wacvblue]{hyperref}

\newcommand{\es}{\text{\o}}

\def\wacvPaperID{733} 
\def\confName{WACV}
\def\confYear{2026}

\title{Zero-Shot Video Deraining with Video Diffusion Models
}

\author{Tuomas Varanka$^{1*}$, Juan Luis Gonzalez$^2$, Hyeongwoo Kim$^3$, Pablo Garrido$^2$, Xu Yao$^2$\\
University of Oulu$^1$ \hspace{5mm} Flawless AI$^2$ \hspace{5mm} Imperial College London$^3$\\
{\tt\small\{juanluis.gonzalez, pablo.garrido, xu.yao\}@flawlessai.com}\\
{\tt\small tuomas.varanka@student.oulu.fi} {\tt\small hyeongwoo.kim@imperial.ac.uk}\\
{\small Project Website: \href{http://tvaranka.github.io/ZSVD}{tvaranka.github.io/ZSVD}}
}

\begin{document}
\maketitle
\begin{abstract}
Existing video deraining methods are often trained on paired datasets, either synthetic, which limits their ability to generalize to real-world rain, or captured by static cameras, which restricts their effectiveness in dynamic scenes with background and camera motion. Furthermore, recent works in fine-tuning diffusion models have shown promising results, but the fine-tuning tends to weaken the generative prior, limiting generalization to unseen cases.
In this paper, we introduce the first zero-shot video deraining method for complex dynamic scenes that does not require synthetic data nor model fine-tuning, by leveraging a pretrained text-to-video diffusion model that demonstrates strong generalization capabilities. 
By inverting an input video into the latent space of diffusion models, its reconstruction process can be intervened and pushed away from the model's concept of rain using negative prompting. 
At the core of our approach is an attention switching
mechanism that we found
is crucial for maintaining dynamic backgrounds as well as structural consistency between the input and the derained video, mitigating artifacts introduced by naive negative prompting. Our approach is validated through extensive experiments on real-world rain datasets, demonstrating substantial improvements over prior methods and showcasing robust generalization without the need for supervised training.
\end{abstract}    
\section{Introduction}
\label{sec:intro}


\begin{figure}[t]
    \small
    \centering
    \setlength{\tabcolsep}{2pt}
    \begin{tabular}{cc}
        \includegraphics[width=0.47\linewidth]{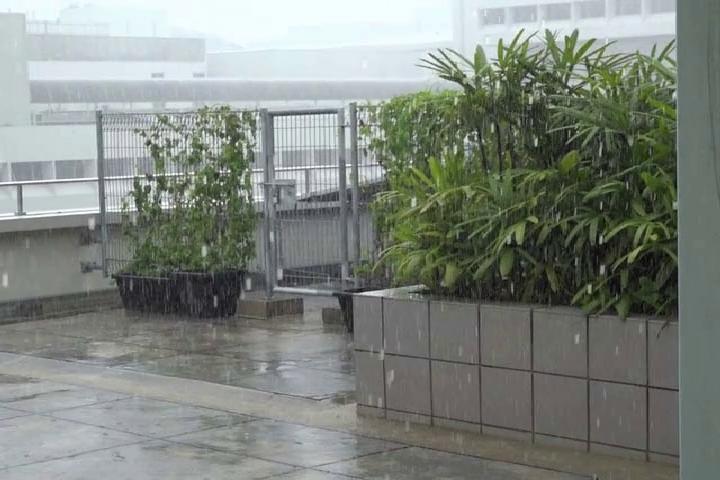} &
        \includegraphics[width=0.47\linewidth]{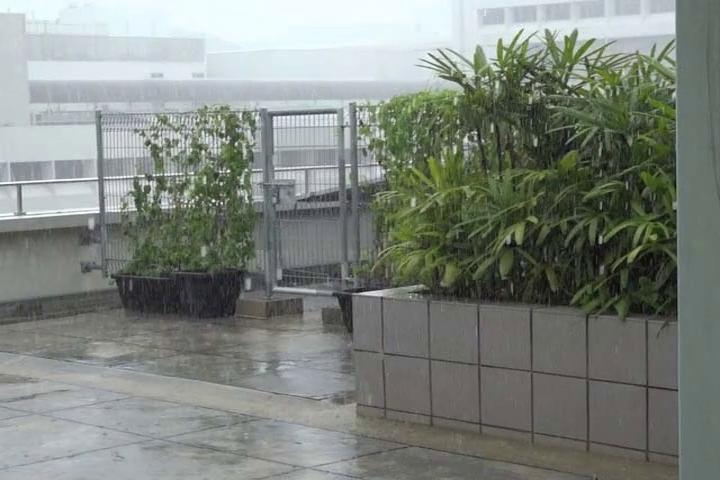}
        \\
        (a) Input & (b) RainMamba~\cite{wu2024-rainmamba} \\
        \includegraphics[width=0.47\linewidth]{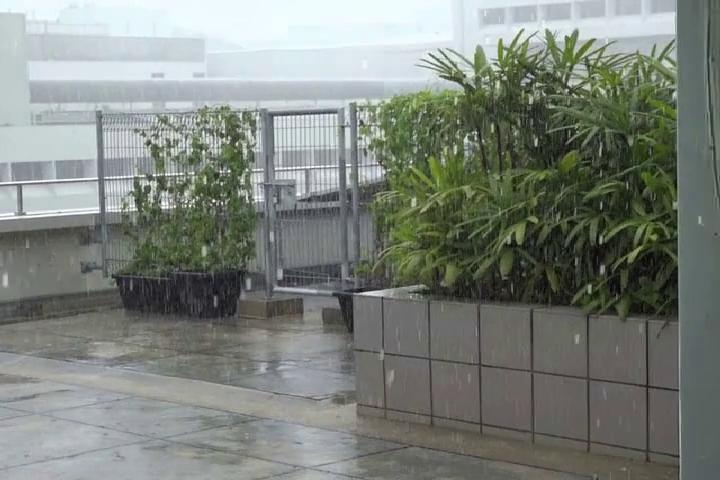} &
       \includegraphics[width=0.47\linewidth]{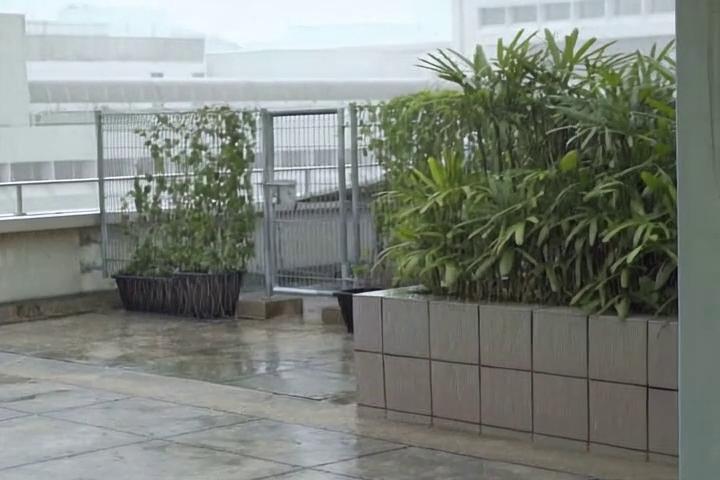} 
       \\
        (c) TURTLE~\cite{ghasemabadi2024-turtle} & (d) Ours
    \end{tabular}
    \caption{
    \textbf{Visual results on video deraining on real rain video.} Compared with state-of-the-art methods such as RainMamba~\cite{wu2024-rainmamba} and TURTLE~\cite{ghasemabadi2024-turtle}, our zero-shot solution is more effective in removing rain streaks and generates temporally consistent results. Please refer to the supplementary for the corresponding video.
    }
    \label{fig:teaser}
\end{figure}

\begin{figure}[t]
    \renewcommand{\arraystretch}{1}
    \small
    \centering
    \setlength{\tabcolsep}{2pt}
        \begin{tabular}{*2c}
            \includegraphics[width=0.47\linewidth]{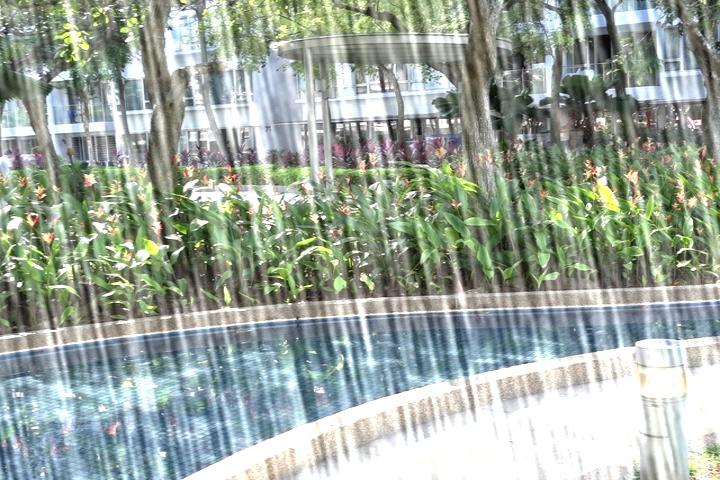} &
            \includegraphics[width=0.47\linewidth]{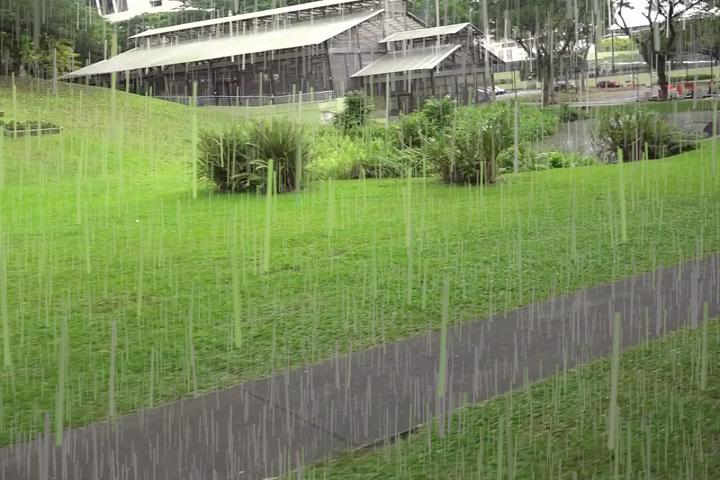}
            \\
            \multicolumn{2}{c}{(a) Synthetic examples from NTURain~\cite{chen2018-ntu_rain}.} 
            \\
            \includegraphics[width=0.47\linewidth]{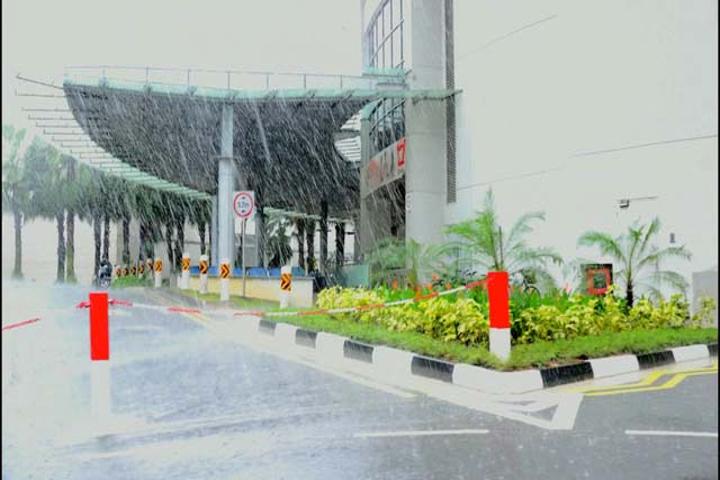} &
            \includegraphics[width=0.47\linewidth]{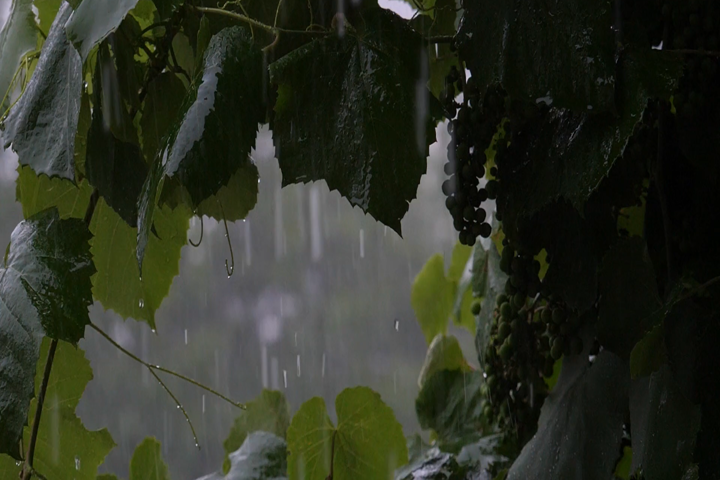}
            \\
            \multicolumn{2}{c}{(b) Real-world examples. } 
            \\
            \includegraphics[width=0.47\linewidth]{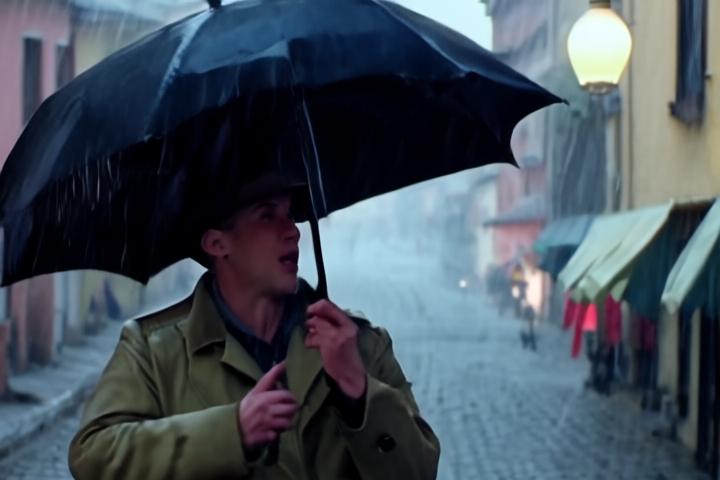} &
            \includegraphics[width=0.47\linewidth]{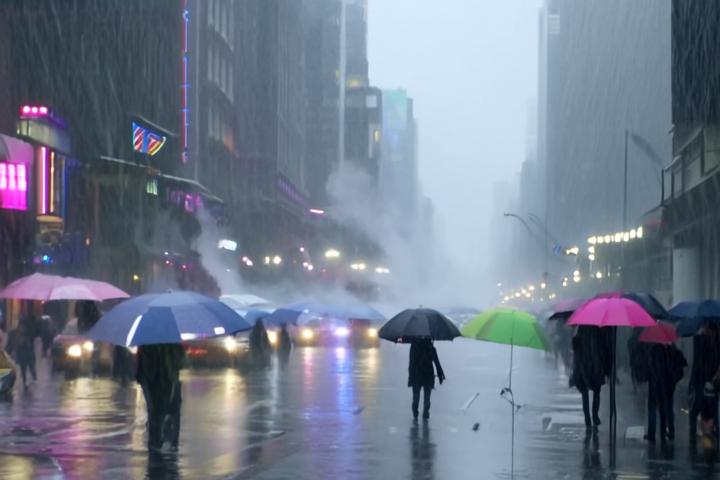}
            \\
            \multicolumn{2}{c}{(c) Generated examples using CogVideoX~\cite{yang2024-cogvideox}.} 
        \end{tabular}
	\caption{
	\textbf{Example cases of rain.} Notably, synthetic rain often exhibits unrealistic repetitive patterns not present in real-world scenes while lacking interactions with the scene objects and depth. Rainy scenes generated by large diffusion models show more realistic effects. Refer to the supplementary for the respective videos.
    }
	\label{fig:rain_examples}
\end{figure}

Rain can severely degrade the quality of captured video scenes due to occluding rain streaks. The degradation not only makes the video content more challenging to view but also adversely affects computer vision tasks, such as tracking and segmentation~\cite{gella2023-weatherproof}. Most existing approaches~\cite{zhang2023-weatherstream, ghasemabadi2024-turtle, liu2024-diff-plugin, wu2024-rainmamba, wei2021-deraincyclegan, wen2024-neural_schroedinger_rain, WeatherWeaver} focus on supervised learning, which requires ground truth data of the same sample with and without rain. As collecting such data is extremely difficult, synthetic data has been frequently used. However, synthesizing realistic rain is challenging due to its complex interactions with the environment, depth variations, and translucency. As a result, methods trained on synthetic data often face a significant domain gap when applied to real-world scenarios. 
\cref{fig:rain_examples} shows examples of synthetic rain (top row) and real rain (middle row). To mitigate or completely avoid this domain gap, semi-supervised and unpaired learning approaches \cite{wei2021-deraincyclegan, wen2024-neural_schroedinger_rain, ye2022-nlcl, wen2024-upim} have been explored.
These approaches, however, frequently fail in reconstructing background details or fully removing rain streaks.

\footnotetext[1]{Work done during an internship at Flawless AI}

Diffusion models have been used in various restoration tasks \cite{liu2024-diff-plugin, tumanyan2023-pnp, lin2023-diffbir, luo2024-daclip} due to their powerful priors of various attributes and concepts learned from large-scale text-to-image training. To adapt the pre-trained diffusion models for restoration tasks, they must be modified to accept images as input, which calls for additional modules and training.
However, introducing and training additional modules alters the original data flow,
which can degrade the robust priors already embedded in the pre-trained models. Moreover, this process relies on synthetically generated data for fine-tuning,
which creates a mismatch with real-world data, diminishing the effectiveness of the generative prior. And although rain exhibits distinct temporal patterns that could be leveraged, most deraining approaches~\cite{liu2024-diff-plugin, wei2021-deraincyclegan, ye2022-nlcl, kim202-nsb, wen2024-neural_schroedinger_rain}
focus solely on the image domain. Real-world rain streaks, being transparent and often small, can easily blend into the background, making it highly ambiguous to discern them from stationary images alone.
Another limitation of image-based diffusion models is that they do not ensure temporal stability and background structure consistency.

To combat the challenges of real-world data and the temporal aspect, we propose a \textit{zero-shot} paradigm for restoration that only leverages a large-scale video diffusion model, eliminating the need for additional training or fine-tuning.
Unlike previous U-Net-based methods that use Stable Diffusion~\cite{woolf2022-stable_diffusion} as their backbone, we opt for a transformer-based video generative model~\cite{yang2024-cogvideox}, which achieves superior video generation quality with high-fidelity motion. To fully exploit the capability of video diffusion models, we utilize inversion~\cite{song2022-ddim} to avoid training of new modules. Inversion finds the latent $x_T$ from which the diffusion model generated a sample, allowing for the reconstruction of the sample $x_0$ from this latent $x_T$. By intervening in the reconstruction process \cite{meng2021-sdedit, hertz2022-prompt2prompt, mokady2023-null_text_inversion}, modifying the condition and applying classifier-free guidance~\cite{ho2021-classifier_free}, we can alter and guide the desired outcome. However, simply removing the rain-related condition during reconstruction is a naive approach and does not effectively remove the rain.

To push the reconstruction path away from a rainy video, we explicitly compute the difference between a reconstruction step with a null condition and a rain condition. This difference is then amplified and added to the reconstructed latent. However, amplifying differences may introduce artifacts, such as background distortions. To address this, we introduce a novel \textit{attention-switching} technique that leverages the $KV$ matrices associated with the null text features in specific transformer blocks. Prior works~\cite{hertz2022-prompt2prompt,cao2023-masactrl,geyer2023-tokenflow} show that cross- and self-attention layers of Stable Diffusion~\cite{rombach2022-stable_diffusion} encode rich semantic information. 
However, manipulating attention in diffusion transformer (DiT) based models~\cite{peebles2023-dit} is challenging, as the $Q$, $K$, and $V$ matrices are computed from a concatenation of vision and text tokens, resulting in joint-attention and entangled features. 
This differs from U-Net-based models~\cite{rombach2022-stable_diffusion,woolf2022-stable_diffusion}, which allow explicit access to cross-attention between text and image.
So far, only a few works \cite{tewel2024-addit, huang2024-scalingconcept} have explored the attention layers in DiT-based models. To handle the entangled features from joint attention, we employ a cross-condition hidden feature that combines text and video features from different conditions. This ensures alignment between the features when modifying the attention process. By manipulating the attention process from a subset of blocks we balance background preservation with deraining ability.

We conduct extensive evaluation experiments across multiple real-world rain datasets. Our approach surpasses state-of-the-art video deraining baselines in both quantitative metrics and qualitative results, achieving superior rain removal and temporal consistency. The contributions of our work can be summarized as follows:
\begin{itemize}
    \item We propose a new \textit{training-free} diffusion paradigm for restoration tasks that circumvents the difficulties in generating realistic synthetic data. To the best of our knowledge, our approach is the first zero-shot method for video deraining. 
    \item We introduce an attention-switching mechanism for transformer-based video diffusion models, designed to preserve scene content and improve video fidelity.
    \item We analyze the effectiveness of different prompts and find that disentanglement via negative prompts plays a key role in achieving high-quality video deraining.
\end{itemize}

\section{Related Work}
\label{sec:relatedwork}
\noindent\textbf{Rain Removal} \quad
Initial approaches attempt to impose handcrafted priors for deraining \cite{gu2017-joint, li2016-rain, kang2011-automatic, luo2015-removing, wang2017-hierarchical}, but often fail in more complex scenes. Paired learning approaches use synthetic rain data with ground truth \cite{yang2019-dual_flow_derain, chen2024-nerd, chen2025-t3_diffweather, sun2025-histoformer, wu2024-rainmamba, liu2024-diff-plugin,sun2023event}. For video deraining, optical flow has been utilized to capture temporal correlations~\cite{yang2019-dual_flow_derain,yang2020self,sun2025semi}. Another line of work~\cite{zhang2025egvd,wang2023unsupervised,fu2024event,ruan2025pre} leverages event cameras to capture high temporal resolution information.

Mamba-based models have been explored in \cite{wu2024-rainmamba,sun2024hybrid,ruan2025pre} for capturing temporal cues of rain. Diff-Plugin \cite{liu2024-diff-plugin} utilizes a pre-trained diffusion model with task-specific modules. TURTLE \cite{ghasemabadi2024-turtle} utilizes a causal history model to retain information from previous frames for various video restoration tasks, including de-raining and de-snowing. A depth prior and a pool of learnable prompts are used by \cite{chen2025-t3_diffweather} for adverse weather restoration problems. Unpaired approaches use a set of rainy and non-rainy samples \cite{wei2021-deraincyclegan, ye2022-nlcl, wen2024-neural_schroedinger_rain, wen2024-upim}. DerainCycleGAN \cite{wei2021-deraincyclegan} modifies the CycleGAN \cite{zhu2017-unpaired_im2im} for rain removal by using an unsupervised attention-guided rain streak extractor. NLCL \cite{ye2022-nlcl} employs contrastive learning to perform deraining with non-local patches. NSB \cite{wen2024-neural_schroedinger_rain} presents a neural Schrödinger bridge \cite{kim202-nsb} for deraining. More recently, WeatherWeaver~\cite{WeatherWeaver} also leverages a video diffusion model for weather synthesis and removal by training SVD~\cite{blattmann2023-stable_video_diffusion} on a curated dataset containing both synthetic and real data. In contrast to methods that rely on synthetic training data with limited generalization to real-world or learning a mapping from scratch via unpaired data, we adopt a \textit{training-free} approach that leverages a large pretrained diffusion prior.

\noindent\textbf{Diffusion-based restoration} \quad
Several approaches \cite{liu2024-diff-plugin, lin2023-diffbir, luo2024-daclip, wang2022-ddnm, cao2024-zvrd} use the strong generative prior from diffusion models to improve results for restoration. DiffBIR \cite{lin2023-diffbir} adds a ControlNet \cite{zhang2023-adding_conditional} to inject image features into a pre-trained T2I diffusion model. Diff-Plugin \cite{liu2024-diff-plugin} trains small modules to extract degradation-specific features. DA-CLIP \cite{luo2024-daclip} trains a module that optimizes the CLIP \cite{radford2021-clip} image embedding, which is integrated into a diffusion model with cross-attention blocks. To use diffusion models, these methods add extra branches that encode image information, requiring model re-training, which results in deteriorating the existing model bias due to the change of input data. To avoid this, zero-shot approaches bypass training altogether. For instance, DDNM \cite{wang2022-ddnm} uses the null-space to perform zero-shot diffusion restoration. We extend the zero-shot capabilities of diffusion models by improving structural preservation, including text as a prior, and extending the approach to video.

\noindent\textbf{Diffusion-based editing} \quad
Recent advances in diffusion models~\cite{ho2020-denoising_diffusion,song2020-score_based} have shown exceptional progress in text-to-image~\cite{rombach2022-stable_diffusion,saharia2022-photorealistic_tex2im,peebles2023-scalable_diffusion} and text-to-video generation~\cite{ho2022-video_diffusion,blattmann2023-stable_video_diffusion,yang2024-cogvideox}. 
Previous studies have explored the editing capacity of the generative diffusion priors~\cite{meng2021-sdedit,hertz2022-prompt2prompt,zhang2023-adding_conditional}. Many of them adopt classifier-free guidance~\cite{ho2021-classifier_free}, a simple technique to boost the sampling quality. Woolf~\etal~\cite{woolf2022-stable_diffusion} further discover that the generative process could be guided better with negative prompts that the model should exclude. Inspired by that, we adopt a negative prompt-based editing approach for rain removal. To edit real scenes within diffusion models, there exist several inversion methods to invert images into the latent space of diffusion models~\cite{song2020-denoising_diffusion,mokady2023-null_text_inversion,huberman2024-ddpm_inversion}. The edit-friendly DDPM inversion~\cite{huberman2024-ddpm_inversion} achieves state-of-the-art results in terms of reconstruction and editing capacity. As such, we adopt this approach for video inversion.

\begin{figure*}[t]
    \centering
    \includegraphics[width=0.99\linewidth]{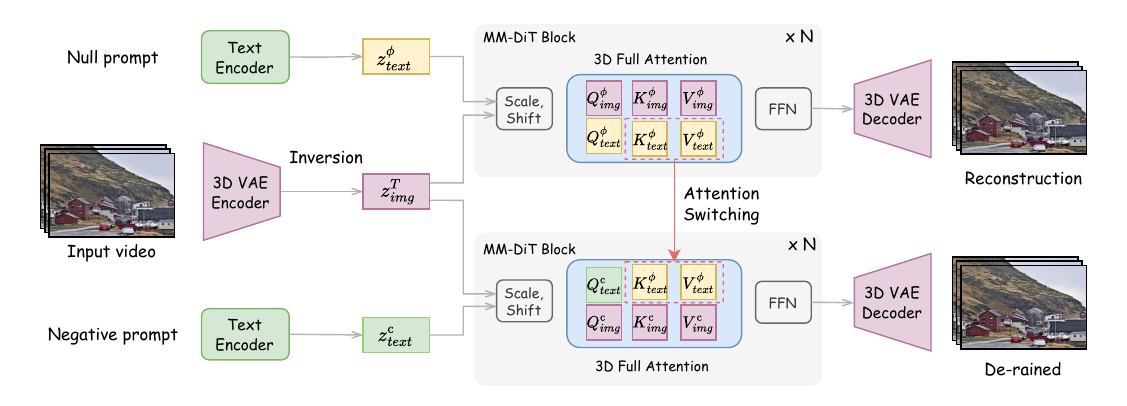}
    \caption{\textbf{Architecture of the proposed zero-shot video deraining approach.} First, video inversion is performed to extract the noise latent $z_{img}^T$. Next, starting from timestep $t_s$ the model performs a reconstruction step with the null prompt and a rain condition step with the negative prompt. The two paths are then combined following \cref{eq:editing}. Attention switching is applied for blocks $\mathcal{B}$, where the $K^{\es}_{text}$ and $V^{\es}_{text}$ are extracted from the null condition and are used to replace their conditional equivalents $K^{c}_{text}$ and $V^{c}_{text}$.
    }
    \label{fig:method}
\end{figure*}
\begin{figure}[t]
\footnotesize
\centering
    \setlength{\tabcolsep}{1pt}
        \begin{tabular}{*3c}
            \includegraphics[width=0.32\linewidth]{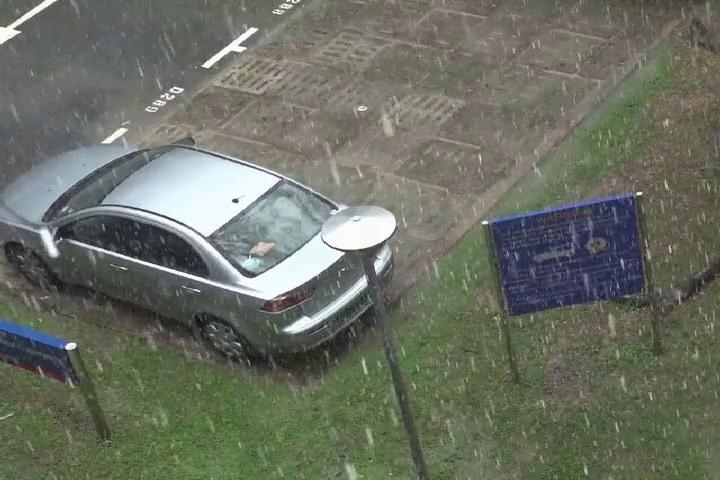} &
            \includegraphics[width=0.32\linewidth]{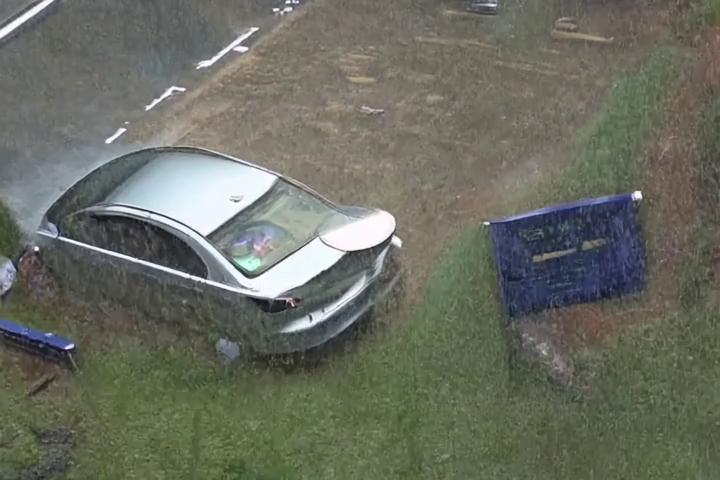} &
            \includegraphics[width=0.32\linewidth]{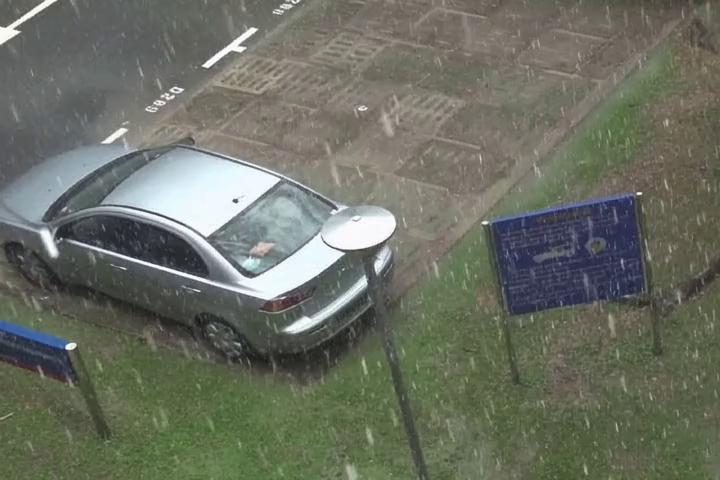}
            \\
            Input & Video DDIM Inversion  & Video DDPM Inversion \\
             & (PSNR = 23.08) & (PSNR = 30.08) \\
        \end{tabular}
	\caption{
	\textbf{Comparison between DDIM~\cite{song2022-ddim} and DDPM~\cite{huberman2024-ddpm_inversion} inversion on video data.} Video DDIM inversion struggles with fully reconstructing the video and misses not only high-frequency details but also larger objects. The PSNR drop in video DDPM inversion is mostly caused by the VideoVAE (PSNR = 31.80) and numerical precision.
	}
	\label{fig:inversion}
\end{figure}

\section{Method}
\label{sec:method}
In this section, we present our method for zero-shot video deraining. Rather than training with synthetic data that generalizes poorly to real-world data, we leverage the powerful generative prior of diffusion models. Our method begins by inverting a given video into the diffusion latent space using an inversion technique. While some earlier inversion methods~\cite{song2022-ddim, mokady2023-null_text_inversion} produce semantic or structural inconsistencies during inversion, recent advancements like DDPM inversion~\cite{huberman2024-ddpm_inversion} can invert images with faithful semantics preservation, making zero-shot restoration possible. By intervening in the reconstruction process, it is then possible to modify the video content, such as degradation effects (e.g., rain), while preserving the underlying structural information and semantics. More specifically, we distill the model's existing knowledge of degradation effect to remove it from the reconstruction process. To further preserve structural information, we propose an \textit{attention-switching} technique that leverages structural information from the $KV$ matrices associated with the joint-attention. The overall pipeline of our proposed method is shown in~\cref{fig:method}.

\subsection{Preliminary}

\paragraph{Video Diffusion}
Text-to-Video diffusion models \cite{ho2022-video_diffusion,blattmann2023-stable_video_diffusion,yang2024-cogvideox} use text input to generate videos from Gaussian noise. Earlier works~\cite{blattmann2023-stable_video_diffusion} adapt text-to-image diffusion models by injecting temporal blocks and fine-tuning on video datasets. These approaches, however, often struggle with temporal consistency and larger motion generation due to the use of 2D VAEs and short-range temporal attention. 

CogVideoX~\cite{yang2024-cogvideox} addresses these issues by using a 3D VAE to compress video data, avoiding temporal inconsistency of individual frame encoding. 

The learning objective remains the same as conditional diffusion~\cite{woolf2022-stable_diffusion}:
\begin{equation}
    L = \mathbb{E}_{x_0, \epsilon \sim \mathcal{N}(0, 1), t, c}\left(||\epsilon - \epsilon_{\theta}(x_t, t ,c)||_2^2\right),
\end{equation}
where $t$ is a timestep, $c$ is the conditional input (text), $\epsilon_\theta$ is the diffusion model, and $x_t$ represents a noisy latent video. To better align the text and image features, the MM-DiT (multimodal diffusion transformer \cite{esser2024-stable_diffusion3}) blocks are used. The hidden features $h = h_{text} | h_{img}$ consist of concatenated text and image features, and the attention is performed simultaneously across both modalities, for better alignment between the modalities.

\noindent\textbf{Diffusion Image Inversion} \quad
Inversion refers to find the latent noise $x_T$ that generates a given sample $x_0$, enabling image editing by manipulating the reconstruction process. SDEdit~\cite{meng2021-sdedit} uses the simplest form of inversion by simply adding Gaussian noise, which limits the preservation of structural information. 
DDIM~\cite{song2022-ddim} further adopts a sampling process that is deterministic, compared to the stochastic DDPM \cite{ho2020-denoising_diffusion}. The DDIM process can hence be inverted by re-arranging the sampling process with respect to the current step $x_t$ and computing the score estimate $\hat{\epsilon}_\theta$ from $x_{t-1}$ instead of $x_{t}$. Several approaches have been proposed to improve diffusion inversion \cite{mokady2023-null_text_inversion, huberman2024-ddpm_inversion, garibi2024-renoise, miyake2023-negative_prompt_inversion, wallace2023-edict, hong2024-dpm_inversion}, with some of them learning additional information like text embeddings \cite{mokady2023-null_text_inversion} or added noise \cite{huberman2024-ddpm_inversion}.

\noindent\textbf{Diffusion Video Inversion} \quad
As no prior work exists on inverting latent videos, we experiment with several options \cite{meng2021-sdedit, song2022-ddim, mokady2023-null_text_inversion, huberman2024-ddpm_inversion}. See \cref{fig:inversion} and the supplementary material for results. Based on these results we utilize DDPM inversion~\cite{huberman2024-ddpm_inversion}, which first gathers the noisy latents $x_t$ from
\begin{equation}
    x_t = \sqrt{\Bar{\alpha}_t}x_0 + \sqrt{1 - \Bar{\alpha}_t}\Tilde{\epsilon}_t,
\end{equation}
associated with each timestep and then computes the noise required for cleaning $x_t$ from the previous $x_{t-1}$ from
\begin{equation}
    z_t = \frac{x_{t-1} - \mu_t(x_t, c)}{\sigma_t}.
\end{equation}
Each of the $z_t$ contains the information required for reconstructing the video at timestep $t$. The video can then be reconstructed from the set ${x_t, z_t, \ldots, z_1}$ by starting from $x_t$, predicting $x_{t-1}$ and adding the extracted noise map $z_t$ progressively, until reaching $x_0$.

\subsection{Training-free Rain Removal}
\label{sec:rain_removal}
Instead of fully reconstructing the video from the inverted latent $x_T$, it is possible to modify the outcome by intersecting the reconstruction process. Here, we aim to modify the video by removing any unwanted degradation effect, i.e., rain, while retaining the background information. A naive approach would simply perform inversion with a rain condition and then remove that condition in reconstruction. However, this does not significantly alter the diffusion path and results in mostly reconstructing the original sample. We find that for sufficient results, the path needs to be explicitly separated from the reconstruction path. We achieve the above via negative prompt editing, akin to \cite{miyake2023-negative_prompt_inversion}.

Pushing the score estimate $\hat{\epsilon}_\theta(x_t)$ away from the degradation effect is done by using an additional term $\epsilon_\theta(x_t) - \epsilon_\theta(x_t, c)$, where $c$ denotes the degradation effect, i.e., rain. Amplifying the difference term with a scalar $\lambda$ and adding the difference to the original score estimate allows us to separate the reconstruction from the degradation effect. This corresponds to the classifier-free guidance term \cite{ho2021-classifier_free} used to guide image generation. Mathematically, the above is performed as follows:
\begin{equation}
    \hat{\epsilon}_\theta(x_t) = \epsilon_\theta(x_t) + \lambda(\epsilon_\theta(x_t) - \epsilon_\theta(x_t, c)).
    \label{eq:editing}
\end{equation}
Note that we can target different levels of rain by varying the trade-off factor $\lambda$. To avoid deviating too much from the reconstruction path, which would result in lower fidelity, a skipping timestep $t_s$ is used. For initial steps, the sample fully follows the reconstruction path; it deviates from it only after $t_s$.

\noindent\textbf{Which rain conditions are effective?}\quad
The condition $c$ in \cref{eq:editing} is a textual prompt, consistent with the use of T2V models. It should only be related to rain to prevent the model from modifying other parts of the sample. We find that the model's ability to disentangle a concept plays a crucial role in the deraining process. 

We first experiment with a simple prompt, \textit{``rain''}, which removes some rain but generally performs poorly. To better understand the reason, we generate a sample using the prompt \textit{``rain''} but did not observe clear rain pattern, indicating it has not been able to disentangle the rain concept.

Next, we conduct a large-scale analysis from $1K$ real-world rainy video crops. These videos are captioned using a video captioner~\cite{hong2024-cogvlm2} and the text-embeddings are extracted using the T5 text-encoder~\cite{raffel2020-t5}. We then extract the text-embeddings associated with the word \textit{``rain''}, compute their mean, and use this as the condition for deraining. When prompting the mean text-embedding, we observe improved performance over the base prompt \textit{rain''}, with a clearer rain footprint in the generated samples. However, some rain streaks and faint background artifacts remain, suggesting that the prompt is not fully disentangled. 

By analyzing the text-embeddings and their respective prompts further, we observe that context from neighboring words in the text-encoder~\cite{raffel2020-t5} plays a crucial part in the disentanglement of a concept. For instance, a simple prompt like \textit{``light rain''} demonstrates better disentanglement of rain from the background and performs better in deraining.

\subsection{Attention Switching in DiTs}\label{sec:block}
\begin{figure}
    \centering
    \includegraphics[width=0.85\linewidth]{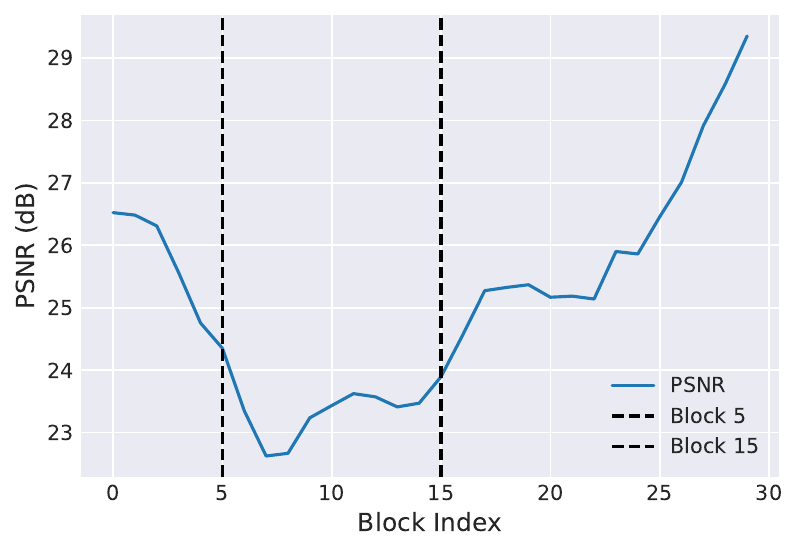}
    \caption{\textbf{Impact of skipping an individual block on the PSNR.} Losing high-frequency information, which is captured by blocks 0-5 and blocks 15-30, has a lower impact on PSNR. 
    }
    \label{fig:blocks_psnr}
\end{figure}

When analyzing the cross-attention maps of the diffusion model, we empirically observe that not all blocks contribute to high-frequency information. To determine whether a block is responsible for low- or high-frequency information, we conduct a study by skipping blocks one-by-one. For each block, we replace the conditional features with null features during generation. We then compare the result to the case with all blocks active in \cref{fig:blocks_psnr}. This experiment is performed using 10 prompts, and we compute the average PSNR impact of each block by averaging its impact across these prompts. It can be observed that the high PSNR values are mainly in the initial blocks 0-5 and later blocks 15-30, indicating they contain high-frequency information.

\noindent\textbf{Attention Switching} \quad
We refer to the selected blocks containing high-frequency information as $\mathcal{B}$. A similar observation on the impacts of different blocks in DiTs is also found in \cite{lv2024-fastercache}. By using another observation made with image diffusion models, where the key $K$ and value $V$ matrices mostly contain the structure of the image \cite{cao2023-masactrl}, we are able to perform stronger deraining with a larger $\lambda$ on the blocks that mostly contain high-frequency information, while simultaneously retaining video fidelity by limiting the impact from blocks $\mathcal{B}$.

We modify the attention block
\begin{equation}
    Attn_b =
    \begin{cases}
      Attn(Q^c, K_{text}^{\es} | K_{img}^c, V_{text}^{\es} | V_{img}^c) & \text{if} \; b \in \mathcal{B}\\
      Attn(Q^c, K^c, V^c) & \text{else},\\
    \end{cases}
    \label{eq:kv_swap}
\end{equation}
by replacing the $K_{text}^c$ and $V_{text}^c$ with their respective ones from the null condition. By using $K_{text}^{\es}$ and $V_{text}^{\es}$ the structural information from the reconstruction latent can be used to better retain content in the video. As only the blocks $\mathcal{B}$ are modified that do not contain high-frequency information, the model's deraining ability is not affected.

However, the use of MM-DiT \cite{esser2024-stable_diffusion3} blocks in CogVideoX \cite{yang2024-cogvideox} presents challenges that restrict the use of \cref{eq:kv_swap}. These issues stem from the joint-attention that limits passing text-embeddings only to the first transformer block instead of all individual blocks as with previous UNet architectures like Stable Diffusion \cite{woolf2022-stable_diffusion}. The conditional text-embeddings serve as input only to the first block and the remaining blocks receive processed hidden states from the previous blocks. Due to the joint-attention the text features are updated and influenced by the visual features, effectively entangling them in the process as information is stored in the text features \cite{toker2025paddinganalysis}.

\noindent\textbf{Cross-condition hidden features}\quad
When applying \cref{eq:kv_swap} directly on MM-DiT blocks without any modifications the output will result in a significant number of artifacts as the newly switched null conditioned text features $h^{\es}_{text}$ do not align with the visual features $h^c_{img}$. This is because the image and text features get entangled during the denoising process as hidden hidden features are passed from one block to the next one. Instead of using the null hidden feature $h^{\es} = h^{\es}_{text} | h^{\es}_{img}$, the cross-condition hidden feature
\begin{equation}
    h^{\es,c} = h^{\es}_{text} | h^{c}_{img},
\end{equation}
is used to obtain the attention inputs $K = P_{K}(h^{\es,c})$ and $V = P_{V}(h^{\es,c})$ to be used in place of the null text inputs in \cref{eq:kv_swap}. This way, the null conditioned matrices $(KV)^{\es}$ for the text will be aligned with the prompt conditioned $(KV)^{c}$ visual features. 

\noindent\textbf{Split attention for retaining conditional text features}
Since the attention switching is only performed for blocks $\mathcal{B}$ this causes a problem due to the forward processing of hidden features. If the key $K$ and value $V$ matrices are changed from conditional to null for the first block following \cref{eq:kv_swap}, it will remain changed for the remaining blocks as the text features are obtained from the previous block. However, we only want this to be the case for blocks $\mathcal{B}$, not all of them. To avoid this, we first compute the hidden features associated with the conditional prompt and discard the visual features using standard attention. 
\begin{equation}
\begin{aligned}
    h^c_{text, b} &= block_i(h^c_{text, b-1}, h^c_{img, b-1}) \\
    h^c_{img, b} &= block_i(h^{\es,c}_{text, b-1}, h^c_{img, b-1})
\end{aligned}
\quad, b \in \mathcal{B}_{\text{initial}}
\end{equation}
Then, we compute the visual features using the attention switching with the cross-condition hidden features. This process ensures that not only are all the features aligned but that the conditional text features are retained through all blocks. This attention split is only necessary for the early blocks $\mathcal{B}_{\text{initial}}$ to ensure that the conditional text features $h^c_{text}$ can be used in the later blocks.

\begin{figure*}[t]
\small
\renewcommand{\arraystretch}{1}
\centering
\setlength{\tabcolsep}{1pt}
    \begin{tabular}{*5c}
        \includegraphics[width=0.19\linewidth]{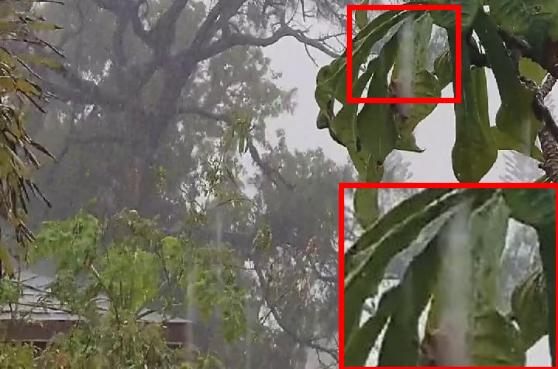} &
        \includegraphics[width=0.19\linewidth]{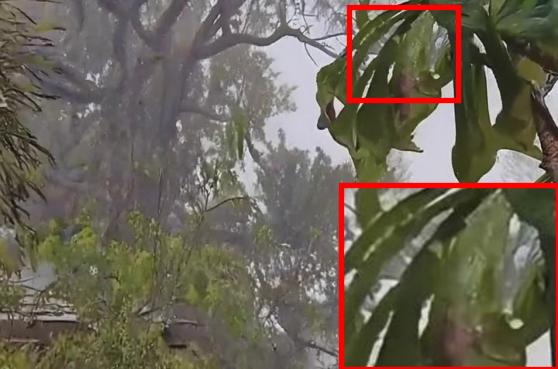} &
        \includegraphics[width=0.19\linewidth]{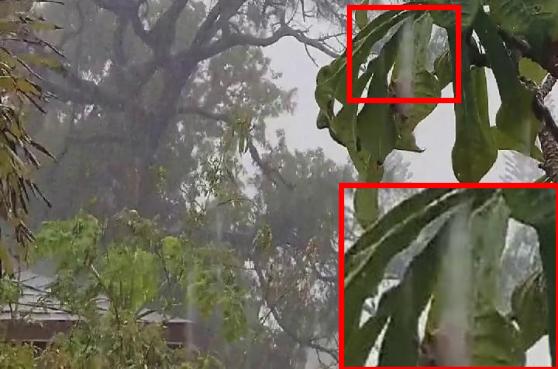} & \includegraphics[width=0.19\linewidth]{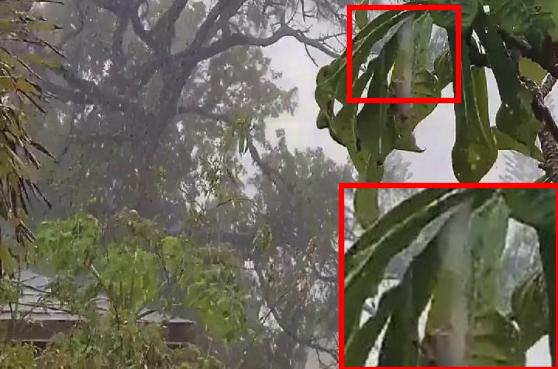} & \includegraphics[width=0.19\linewidth]{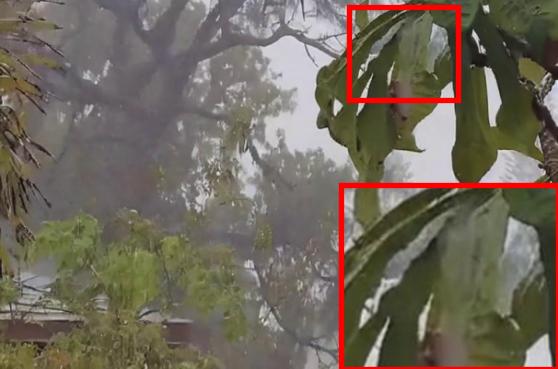}
        \\
        \includegraphics[width=0.19\linewidth]{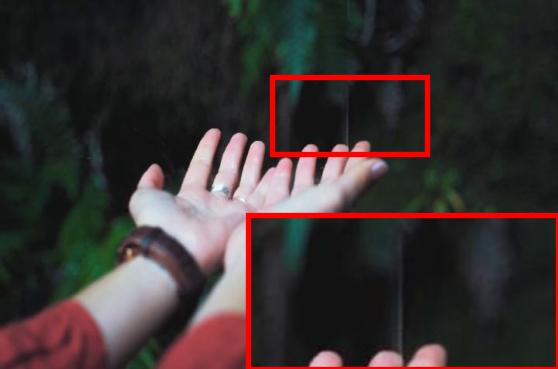} &
        \includegraphics[width=0.19\linewidth]{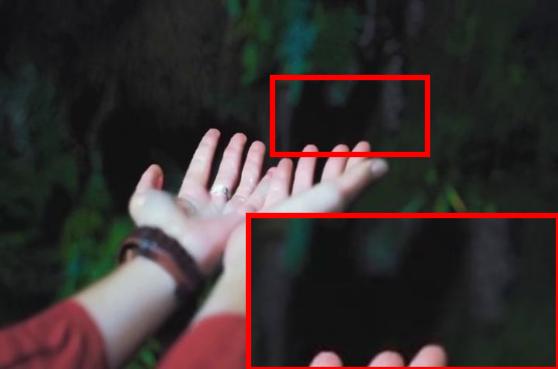} &
        \includegraphics[width=0.19\linewidth]{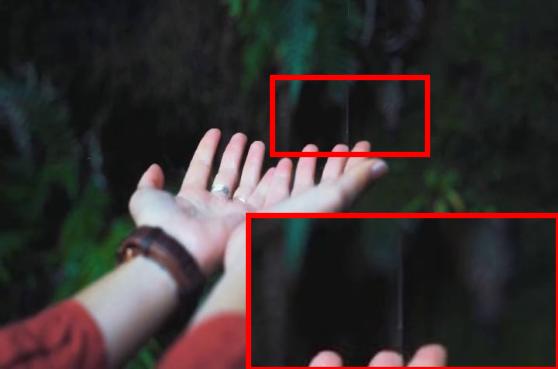} & \includegraphics[width=0.19\linewidth]{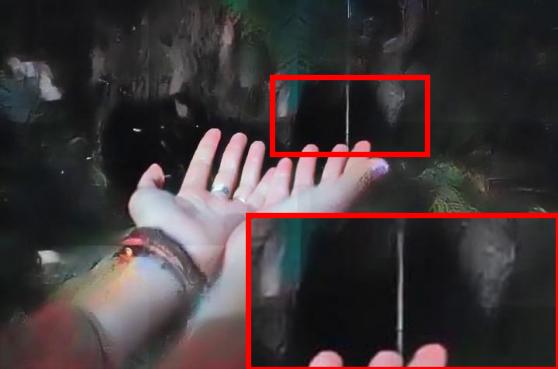} & \includegraphics[width=0.19\linewidth]{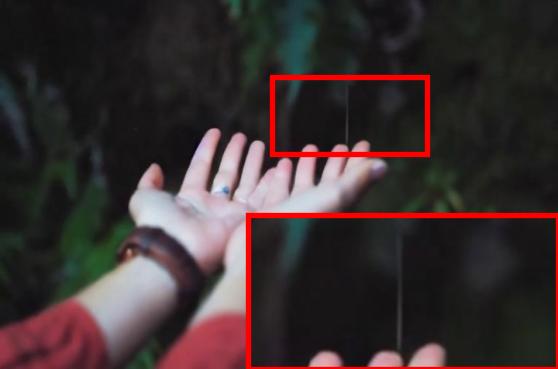}
        \\
        \includegraphics[width=0.19\linewidth]{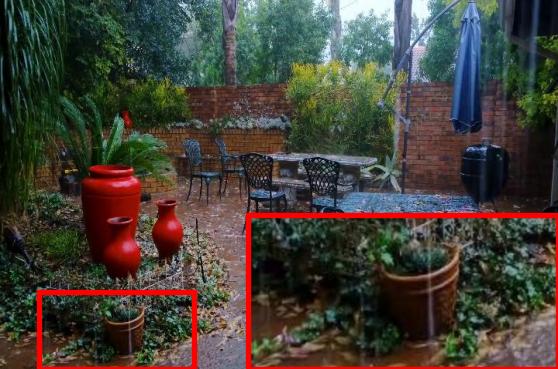} &
        \includegraphics[width=0.19\linewidth]{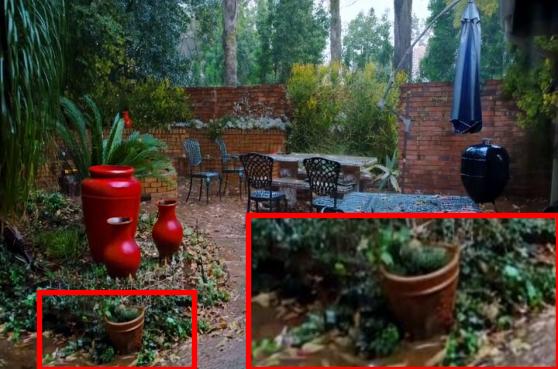} &
        \includegraphics[width=0.19\linewidth]{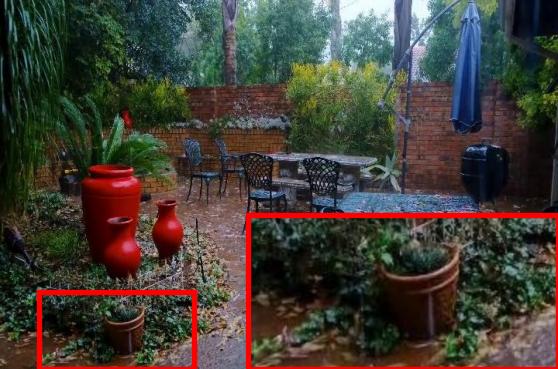} & \includegraphics[width=0.19\linewidth]{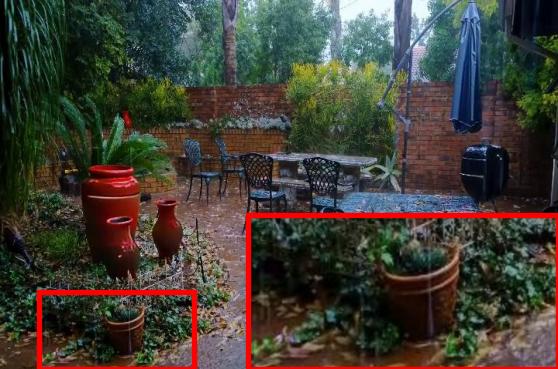} & \includegraphics[width=0.19\linewidth]{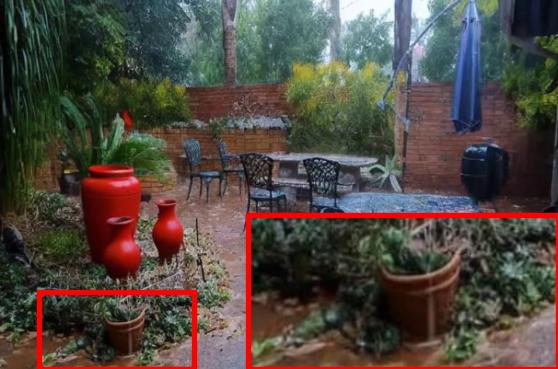}
        \\
        Input & Ours & RainMamba~\cite{wu2024-rainmamba} & Turtle~\cite{ghasemabadi2024-turtle} & Diff-Plugin~\cite{liu2024-diff-plugin} \\
    \end{tabular}
\caption{
\textbf{Qualitative comparison of deraining on real-world videos using different approaches.} Our method achieves effective rain removal, faithfully preserving the original content and temporal consistency. Please refer to the supplementary for best viewing experience.
}
\label{fig:results}
\end{figure*}
\begin{figure}[h]
    \centering
    \includegraphics[width=0.9\linewidth]{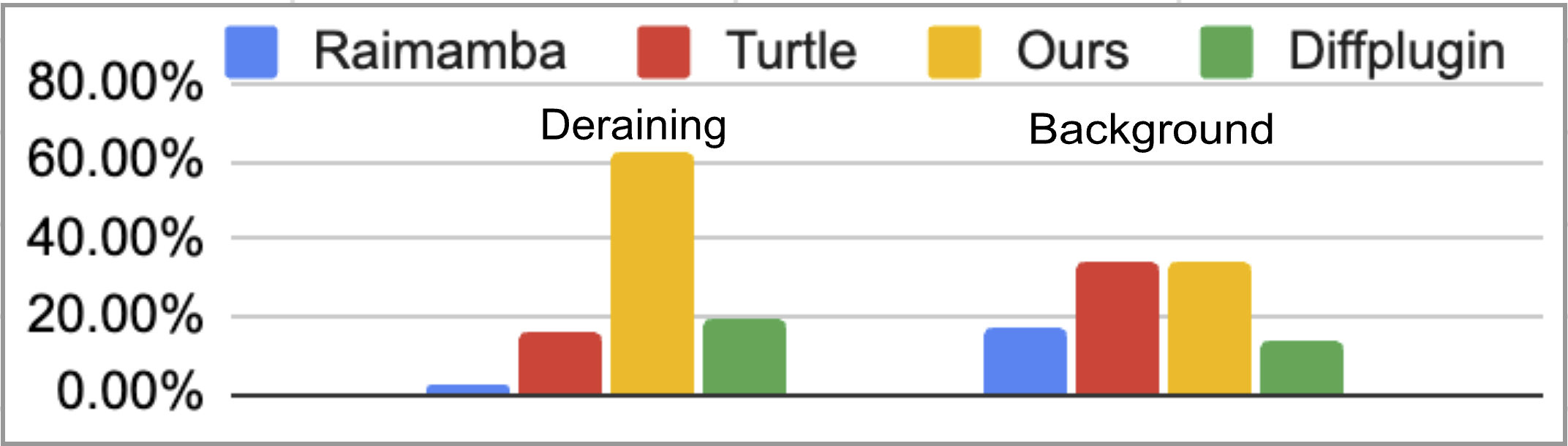}
    \\
    \includegraphics[width=0.9\linewidth]{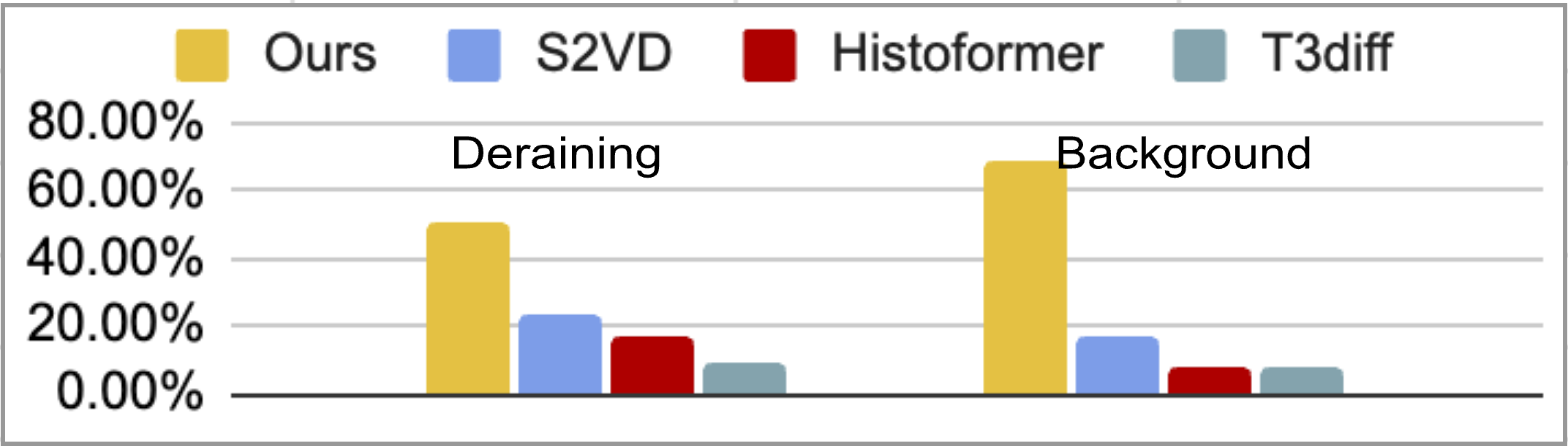}
    \caption{\textbf{User study.} Top: Comparison with RainMamba \cite{wu2024-rainmamba}, Turtle \cite{ghasemabadi2024-turtle}, Diff-Plugin \cite{liu2024-diff-plugin} and ours. Bottom: Comparison with S2VD \cite{yue2021-s2vd}, Histoformer \cite{sun2025-histoformer}, T3DiffWeather \cite{chen2025-t3_diffweather} and ours. 
    }
    \label{fig:userstudy}
\end{figure}


\section{Experiments and Results}
\label{sec:results}

\paragraph{Experimental settings}
In our experiments, We employ the 2B-variant of CogVideoX~\cite{yang2024-cogvideox}. Regarding the hyperparameter $\lambda$, we observe that the optimal value varies depending on the rain intensity. For simplicity, we set $\lambda = 15$, which performs well for most cases. For the skipping timestep $t_s$, we use $t_s = 40$, although this can be further optimized for individual videos. For the blocks $\mathcal{B}$ used for attention switching, we select the initial blocks $[0-4]$ and later blocks $[15-29]$, based on our block analysis of high frequency information. All the input videos are cropped and resized to a resolution of $720 \times 480$. The number of inference steps is set up to $100$, and the entire deraining process takes approximately 3 minutes on one NVIDIA H100 GPU. 

\noindent\textbf{Datasets.} \quad
Most existing rain datasets are synthetic, as capturing real-world rain data with corresponding deraining ground truth is challenging. NTURain~\cite{chen2018-ntu_rain} includes a real-world rain testing dataset without ground truth. GT-Rain~\cite{ba2022-gt_rain} is a large-scale dataset of real-world rainy video and clean image pairs captured using webcams. However, many low-resolution videos exhibit significant compression artifacts. To address this, we exclude all videos with resolutions below $450 \times 300$. 
To address the need for high-quality real-world rain videos with dynamic scenes, we collected an additional dataset of 13 videos, denoted as \textit{RealRain13}. For experimental validation, we utilize the real-world testing data from NTURain~\cite{chen2018-ntu_rain}, the filtered subset of GT-Rain \cite{ba2022-gt_rain}, and our collected \textit{RealRain13} dataset. This combination ensures a robust evaluation across diverse real-world scenarios.

\noindent\textbf{Evaluation Metrics.} \quad
Evaluating our approach and baselines on real-world rain videos presents some challenges as no ground truth data is generally available. To evaluate the effectiveness of rain removal and the quality of the video, we utilize three metrics adapted to our particular problem: \textit{MUSIQ} \cite{ke2021-musiq}, \textit{CLIP-IQA} \cite{wang2023-clipiqa}, and \textit{Warp Error}.
For video quality assessment, we use a non-reference-based metric, MUSIQ \cite{ke2021-musiq}. As MUSIQ is a per-frame metric, we average out results across video frames to obtain the final score.
For measuring rain removal, we utilize CLIP-IQA, which computes the cosine similarity to text prompts. We modify the metric for rain removal by using a matching prompt.
Warp error measures temporal consistency by the difference between a frame and its previous frame warped using the estimated optical flow~\cite{teed2020raft}. Inconsistent rain removal often leads to flickering in the output video, resulting in higher warp error. 

\noindent\textbf{Comparison baselines.}
We compare our method with several state-of-the-art methods for deraining, including RainMamba \cite{wu2024-rainmamba}, TURTLE \cite{ghasemabadi2024-turtle}, Diff-Plugin \cite{liu2024-diff-plugin}, T3-DiffWeather \cite{chen2025-t3_diffweather}, S2VD \cite{yue2021-s2vd} and HistoFormer \cite{sun2025-histoformer}. For all the methods, we use their official implementations.

\subsection{Qualitative Results}
\cref{fig:results} presents qualitative results on RealRain13 datasets. From \cref{fig:results}, we observe that TURTLE \cite{ghasemabadi2024-turtle} often fails to remove rain and can introduce artifacts (middle row). Diff-Plugin \cite{liu2024-diff-plugin} demonstrates improved rain removal but struggles with retaining details and temporal consistency, which is likely caused by randomness in the diffusion and applying it frame-by-frame. Furthermore, some of the perceived rain removal results from the loss of high-frequency details rather than effective deraining. While RainMamba \cite{wu2024-rainmamba} can remove larger rain streaks, it often fails to remove smaller streaks further in depth. Our proposed method is able to effectively remove rain from all scenes with good temporal consistency and fidelity to the original video. 

We conducted a user study to assess the perceptual quality of the deraining results, shown in \cref{fig:userstudy}. The study was conducted with 24 participants, who were asked to select their preferences among four different methods based on two criteria: deraining quality and background preservation. The first set compared our method with RainMamba~\cite{wu2024-rainmamba}, Turtle~\cite{ghasemabadi2024-turtle} and Diff-Plugin~\cite{liu2024-diff-plugin}, while the second set compared our method with S2VD~\cite{yue2021-s2vd}, Histoformer~\cite{sun2025-histoformer}, and T3diff~\cite{chen2025-t3_diffweather}. Each participant evaluated both sets, covering 34 comparisons, totaling 136 videos and 1,632 answers. 
Our method was the most preferred for deraining quality across all comparisons, and for background preservation in the second set.

\begin{table*}[t]
    \centering
    \small
    \resizebox{0.9\linewidth}{!}{%
      \begin{tabular}{lcccccccccc}
        \toprule[1pt]
        \multirow{2}[1]{*}{Method} &\multirow{2}[1]{*}{\shortstack{Zero- \\ shot}}  & \multicolumn{3}{c}{\textbf{NTURain}} & \multicolumn{3}{c}{\textbf{RealRain13}} & \multicolumn{3}{c}{\textbf{GT-Rain}} \\ \cmidrule[0.5pt](lr){3-5} \cmidrule[0.5pt](lr){6-8} \cmidrule[0.5pt](lr){9-11}
        & & \!\!CLIP-IQA$\downarrow$\!\! & \!\!Warp$\downarrow$\!\! & \!\!MUSIQ$\uparrow$\!\! & \!\!CLIP-IQA$\downarrow$\!\! & \!\!Warp$\downarrow$\!\! & \!\!MUSIQ$\uparrow$\!\! & \!\!CLIP-IQA$\downarrow$\!\! & \!\!Warp$\downarrow$\!\! & \!\!MUSIQ$\uparrow$\!\! \\
        \midrule
        Input & & 0.879 & 0.096 & \underline{59.17} & 0.683 & 0.035 & \textbf{60.12} & 0.133 & 0.0109 & 51.14 \\
        \midrule
        HistoFormer~\cite{sun2025-histoformer} & & 0.742 & 0.059 & 57.06 & 0.632 & 0.023 & 58.54 & 0.133 & 0.0082 & 51.37 \\
        T3-DiffWeather~\cite{chen2025-t3_diffweather} & & 0.659 & 0.057 & 57.92 & 0.574 & 0.023 & 58.50 & 0.124 & 0.0094 & \underline{51.58} \\
        S2VD~\cite{yue2021-s2vd} & & 0.700 & 0.058 & 57.06 & 0.612 & \underline{0.020} & 58.31 & 0.129 & \underline{0.0080} & 51.21 \\
        Diff-Plugin~\cite{liu2024-diff-plugin} & & \underline{0.430} & 0.063 & 51.47 & \underline{0.467} & 0.026 & 55.85 & 0.134 & 0.0113 & 48.59 \\
        RainMamba~\cite{wu2024-rainmamba} & & 0.734 & \underline{0.057} & 56.84 & 0.619 & 0.022 & 58.57 & 0.170 & 0.0097 & \textbf{52.88} \\
        Turtle~\cite{ghasemabadi2024-turtle} & & 0.829 & 0.064 & \textbf{59.82} & 0.595 & 0.027 & \underline{59.94} & \underline{0.116} & 0.0098 & 51.71 \\
        \midrule
        \textbf{Ours} & \checkmark & \textbf{0.324} & \textbf{0.047} & 55.43 & \textbf{0.447} & \textbf{0.015} & 57.00 & \textbf{0.075} & \textbf{0.0064} & 50.18 \\
        \bottomrule[1pt]
      \end{tabular}}
    \caption{\textbf{Quantitative comparisons with state-of-the-art video deraining methods} on NTURain~\cite{chen2018-ntu_rain}, RealRain13, and GT-Rain~\cite{ba2022-gt_rain}.}
    \label{tab:quantitative}
\end{table*}
\subsection{Quantitative Results}
\cref{tab:quantitative} presents the quantitative comparison results on NTURain \cite{chen2018-ntu_rain}, RealRain13 and GT-Rain \cite{ba2022-gt_rain}. Among the evaluated methods, our approach achieves the best result for CLIP-IQA across all datasets. While other methods perform significantly worse, Diff-Plugin attains comparable results. Notably, the input video achieves the best MUSIQ score, likely because MUSIQ is trained on multi-scale image datasets that typically contain no rain as a degradation effect. This metric highlights a key limitation of Diff-Plugin: it frequently alters the background and introduces artifacts. Furthermore, the Warp Error highlights temporal inconsistencies in many compared methods, with our method achieving the lowest error across all datasets, demonstrating better temporal consistency.


\begin{table}[!ht]
    \small
    \centering
    \resizebox{0.95\linewidth}{!}{%
    \setlength{\tabcolsep}{2pt}
    \begin{tabular}{lccccc}
        \toprule
         & \multirow{2}{*}{Ours} & \multicolumn{2}{c}{\textbf{NTURain}} & \multicolumn{2}{c}{\textbf{RealRain13}} \\
        \cmidrule[0.5pt](lr){3-4}\cmidrule[0.5pt](lr){5-6}
         & & CLIP-IQA $\downarrow$ & MUSIQ $\uparrow$ & CLIP-IQA $\downarrow$ & MUSIQ $\uparrow$ \\
        \midrule
        Input & & 0.842 & \textbf{59.17} & 0.652 & \textbf{60.12} \\
        \midrule
        DDIM inv.~\cite{song2022-ddim} & & 0.528 & 52.67 & 0.555 & 57.12 \\
        DDPM inv.~\cite{huberman2024-ddpm_inversion} & \checkmark & 0.324 & 55.43 & 0.447 & 57.00 \\
        \midrule
        Implicit prompt &  & 0.787 & 54.16 & 0.700 & 58.75 \\
        \textit{``Rain''} & & 0.686 & 53.59 & 0.668 & 57.83 \\
        Mean prompts & & 0.525 & 55.13 & 0.601 & 58.74 \\
        \textit{``Light rain''} & \checkmark & 0.324 & 55.43 & 0.447 & 57.00 \\
        \midrule
        w/o attn-switch & & 0.437 & 54.18 & 0.457 & 57.14 \\
        w/ attn-switch & \checkmark & \textbf{0.324} & 55.43 & \textbf{0.447} & 57.00 \\
        \bottomrule
    \end{tabular}%
    }
        \caption{\textbf{Ablation study} on NTURain \cite{chen2018-ntu_rain} and RealRain13.}
    \label{tab:quantitative_ablation}
\end{table}

\subsection{Ablation Study}
We ablate the core components of the approach in \cref{tab:quantitative_ablation}, including video inversion, rain condition and the use of attention-switching technique.

\noindent\textbf{Video Inversion} \quad
We compare the used video DDPM inversion with video DDIM inversion. Since the DDIM inversion performs poorly in editing, set  $t_s = 70$ to avoid large deviations from the scene. From \cref{tab:quantitative_ablation} (top) we can see that The DDIM inversion approach obtains a reasonable performance in CLIP-IQA, but obtains poor result in MUSIQ, indicating poor reconstruction. The deraining ability is weakened by the high $t_s$, and despite it the method performs reconstruction poorly. Compared to image DDIM inversion, the video latent is significantly more complex and has more potential paths. Therefore, a more precise inversion such as the video DDPM inversion is required.
\begin{figure}[t]
\small
\vspace{-3mm}
\centering
    \setlength{\tabcolsep}{1pt}
        \begin{tabular}{*3c}
            \includegraphics[width=0.16\textwidth]{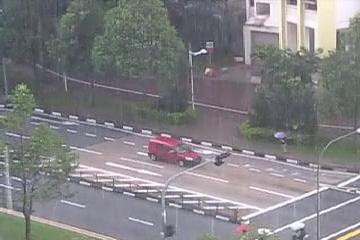} &
            \includegraphics[width=0.16\textwidth]{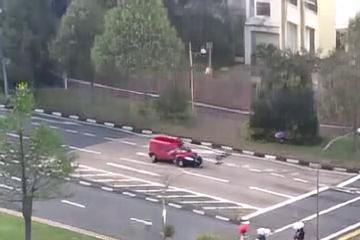} &
            \includegraphics[width=0.16\textwidth]{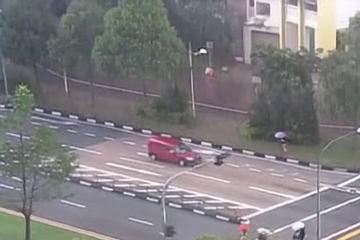}
            \\ 
            Input & Base & Attention switching \\
        \end{tabular}
	\caption{
	\textbf{Ablation study of attention switching.} Although both cases can remove the rain well, the base model struggles with artifacts, such as missing objects, color shifts, and saturation.
	}
        \vspace{-3mm}
	\label{fig:ablation_blocks}
\end{figure}

\noindent\textbf{Prompts} \quad
We evaluate the method with four different prompts. The implicit prompt refers to first inverting the video using \textit{``rain''} and then using an empty prompt during deraining. \textit{``Rain''}, mean of rain prompts and \textit{``light rain''} correspond to the ones used in \cref{sec:rain_removal}. We can observe the CLIP-IQA results in the worst performance with the implicit prompt out of the four prompts. As expected, an explicit push from the rainy concept is required, which can be seen from the improved CLIP-IQA, while MUSIQ remains relatively stable with minor drop for \textit{``rain''} due to the poor disentanglement.

\noindent\textbf{Attention Switching} \quad
When switching the $KV$ matrices in the attention blocks a larger $\lambda = 25$ can be used which results in better deraining, which can be observed from the CLIP-IQA. Simultaneously, the fidelity of the original video is preserved, as observed from the MUSIQ score for NTURain and the visual comparison in~\cref{fig:ablation_blocks}.

\subsection{Discussion and Limitations}
The proposed method relies on the performance of existing video diffusion models. The used CogVideoX \cite{yang2024-cogvideox} is the first open-sourced video diffusion model to use latent videos, instead of latent frames, significantly contributing to the temporal consistency. With the rapid advancement of video diffusion \cite{blattmann2023-stable_video_diffusion, singer2022-make_a_video, menapace2024-snap_video, yang2024-cogvideox, brook2024-sora}, the performance of the proposed approach is expected to improve further. The method struggles with heavy rain, where the rain is constant, as it may be misinterpreted as part of the scene. In this work we focus on deraining, but show in the supplementary that the approach can be generalized to other restoration tasks such as snowing, however, currently it is limited by the performance of the underlying video diffusion model.

\section{Conclusion}
\label{sec:conclusion}
We present a new training-free paradigm for video restoration, focusing on rain degradations, which are hard to synthesize. To circumvent the use of synthetic rain data, which fails to generalize to real-world cases, we leverage a large generative prior from diffusion models. We avoid costly training of additional modules for specific input samples via inversion. We analyze what makes a prompt effective and propose a method for improving sample fidelity by smartly steering the attention process. By utilizing attention features from the reconstruction process and integrating them with attention switching the method improves background preservation. We showcase the method's ability to remove rain in challenging real-world rain scenarios. This work lays the groundwork for addressing different video restoration problems by leveraging large video diffusion models.

\noindent\textbf{Acknowledgements} \quad

\noindent Finnish Foundation for Technology Promotion.

\clearpage
\appendix
\setcounter{page}{1}
\setcounter{figure}{8}
\maketitlesupplementary

Video results for all corresponding figures are organized in \texttt{\small{videos.html}}. Benchmark results on NTURain~\cite{chen2018-ntu_rain}, GT-Rain~\cite{ba2022-gt_rain} and RealRain13 datasets can be found in \texttt{\small{videos/benchmark\_videos}}. \textit{For the best viewing experience, we strongly recommend opening \texttt{\small{videos.html}} alongside this document.}

\section{Implementation Details}
Our backbone model used in all of our experiments is CogVideoX-2b~\cite{yang2024-cogvideox}, a large-scale text-to-video generation model based on a diffusion transformer architecture. We remark that the existing 2B variant model is restricted to precisely 49 frames at a $480 \times 720$ resolution. However, such a limitation has already been addressed in the CogVideoX1.5-5B variant~\footnote{\url{https://huggingface.co/THUDM/CogVideoX1.5-5B-SAT}} and newer releases of diffusion models are likely to improve further. Due to the large size of the diffusion model, the inference time of our method is approximately 2 minutes and 50 seconds on one NVIDIA A100 GPU, with half of the time being allocated to video inversion and the other half to video reconstruction.

\begin{figure}[t]
\footnotesize
\centering
    \setlength{\tabcolsep}{1pt}
        \begin{subfigure}{0.98\linewidth}
        \hspace{-0.2cm}
        \begin{tabular}{*3c}
            \raisebox{35pt}[0.04\linewidth][0pt]{\parbox[t]{3mm}{\rotatebox[origin=c]{90}{\textit{``rain''}}} } &
            \includegraphics[width=0.49\linewidth]{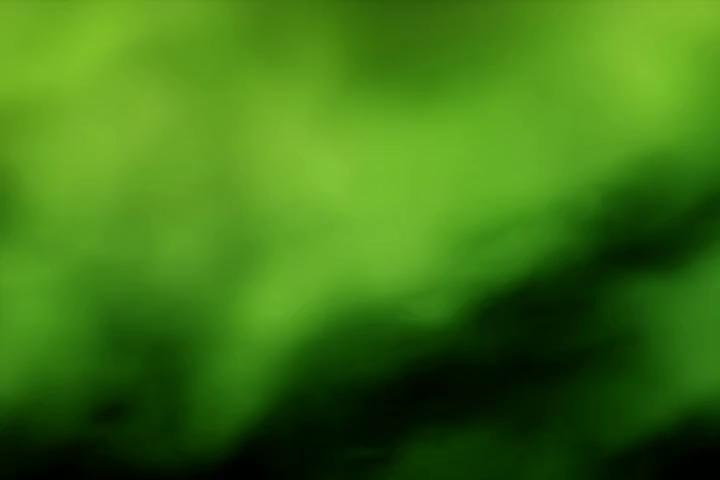} & \includegraphics[width=0.49\linewidth]{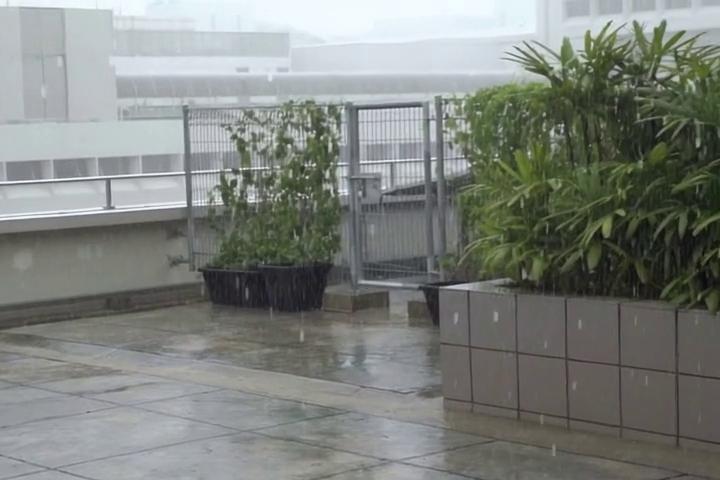}  \\
            \raisebox{35pt}[0.04\linewidth][0pt]{\parbox[t]{3mm}{\rotatebox[origin=c]{90}{Mean of rain prompt}}} &
            \includegraphics[width=0.49\linewidth]{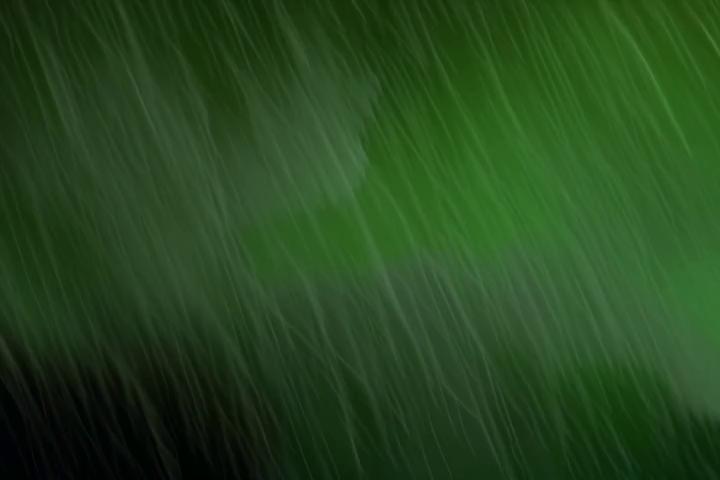} &  
            \includegraphics[width=0.49\linewidth]{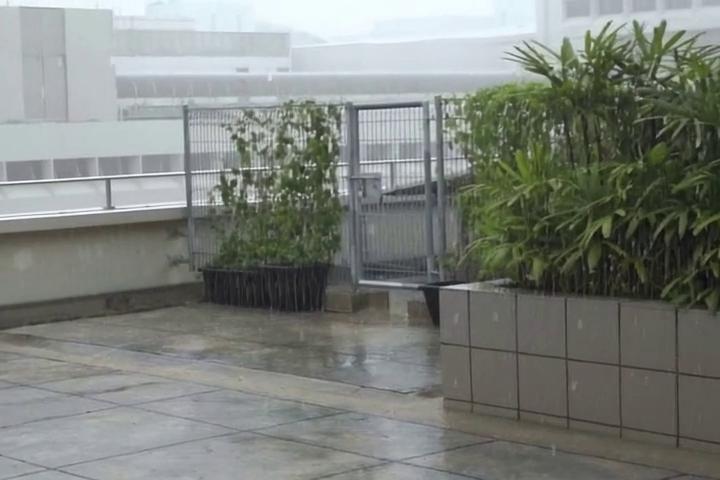}\\
            \raisebox{35pt}[0.04\linewidth][0pt]{\parbox[t]{3mm}{\rotatebox[origin=c]{90}{\textit{``light rain''}}}} &  
            \includegraphics[width=0.49\linewidth]{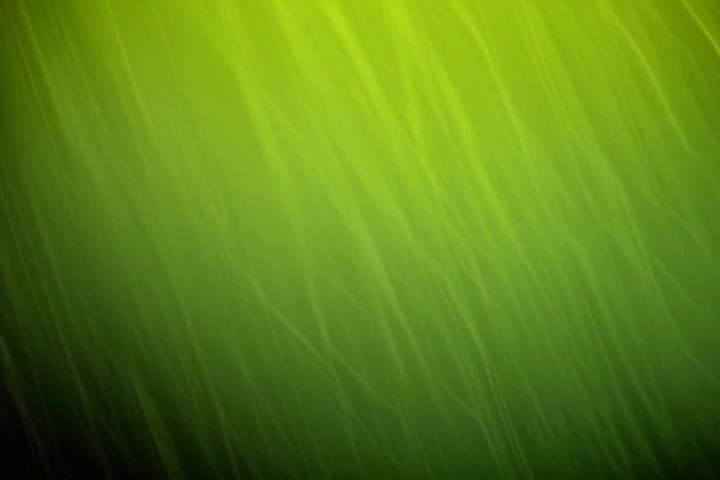} &
            
            \includegraphics[width=0.49\linewidth]{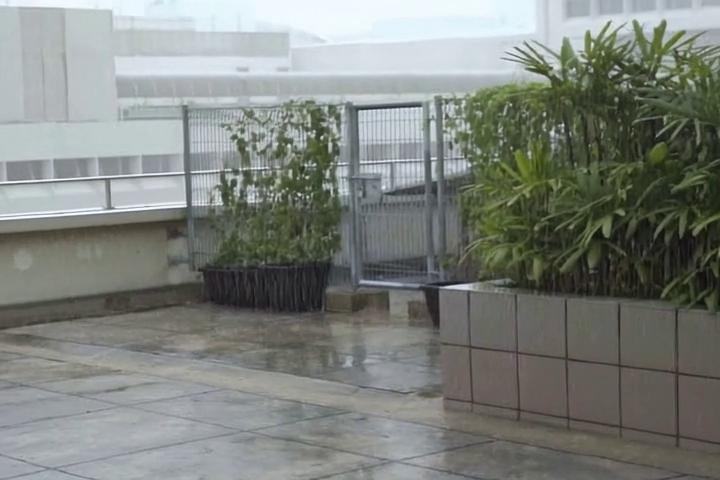}
            \\
             & Generated & Deraining \\
        \end{tabular}
        \end{subfigure}%
	\caption{
	Different rain prompts and their respective results. \textit{Left:} Using diffusion model to generate a video based on the prompt. \textit{Right:} Deraining results with different prompts. The generated video from \textit{``rain''} shows no rain. When using mean of rain prompts, the generated video shows a clear rain pattern and some background, indicating the prompt is not fully disentangled. \textit{``light rain''} shows excellent rain-background disentanglement, and overall it performs the best for deraining.
	}
	\label{fig:rain_prompt}
\end{figure}
\begin{table*}[h]
    \centering
    \setlength{\tabcolsep}{2pt}
    \begin{tabular}{lcc|ccc}
        \toprule
        Method & S2VD & RainMamba & Histoformer & Diff-Plugin & Ours \\
        \midrule
        Trained on NTU & \checkmark & \checkmark & & & \\
        \midrule
        PSNR $\uparrow$ & 37.37 & 37.87 & 29.96 & 24.91 & 27.66 \\
        SSIM $\uparrow$ & 0.9683 & 0.9738 & 0.9112 & 0.7683 & 0.8492 \\
        \bottomrule
    \end{tabular}
    \caption{Quantitative comparisons on NTURain~\cite{chen2018-ntu_rain} synthetic set.}
    \label{tab:ntu_synthetic}
\end{table*}

\section{Rain prompt analysis}
In Sec.3.2 of the main paper, we did an analysis on the rain conditions. Here we present the generated videos corresponding the different rain conditions and the corresponding deraining results in \cref{fig:rain_prompt}.

We first experiment with a simple prompt \textit{``rain''}, which is able to remove some rain but generally performs poorly, see \cref{fig:rain_prompt} (top right). To better understand the reason and analyze the failure, we generate a sample using the prompt, see \cref{fig:rain_prompt} (top left). The prompt shows no visible rain pattern, indicating it has not been able to disentangle the rain concept properly.

Next, we conduct a large-scale analysis from $1K$ real-world rainy video crops. These videos are captioned using an automatic video captioner~\cite{hong2024-cogvlm2} and the text-embeddings are extracted using the T5 text-encoder~\cite{raffel2020-t5}. We then extract the text-embeddings associated with the word \textit{``rain''}, compute their mean, and use this as the condition for deraining. In \cref{fig:rain_prompt} (middle), we show results when prompting the mean text-embedding, computed from the extracted text-embeddings. Such an approach improves over the base prompt \textit{``rain''} and shows a clear rain footprint when used in generation. However, some rain streaks remain, and a faint background can still be observed in the generated sample. We hypothesize that the prompt is not fully disentangled, causing limited deraining performance.
 
In analyzing the extracted text-embeddings and their respective prompts further, we observe that context from neighboring words in the text-encoder \cite{raffel2020-t5} plays a crucial part in the disentanglement of a concept. In \cref{fig:rain_prompt} (bottom), we utilize a simple prompt \textit{``light rain''}, which is able to disentangle rain from the background in the generation and performs best in deraining.

\begin{figure*}[t]
\small
\centering
\setlength{\tabcolsep}{1pt}
\begin{subfigure}{0.99\textwidth}
\hspace{-0.2cm}
    \begin{tabular}{*4c}
        \includegraphics[width=0.25\textwidth]{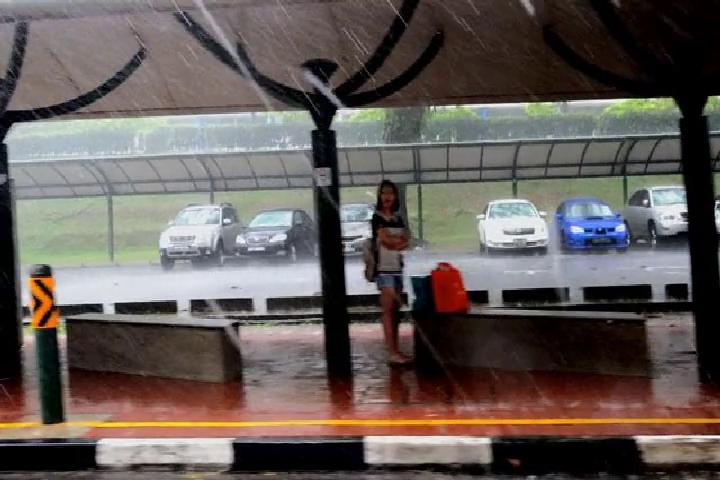} &
        \includegraphics[width=0.25\textwidth]{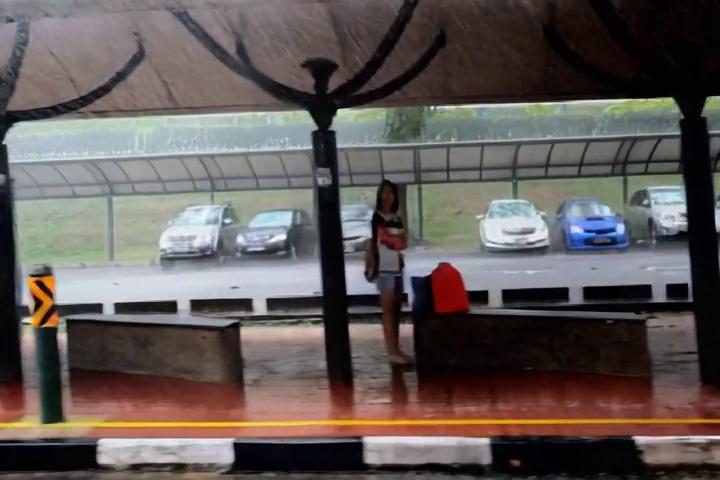} &
        \includegraphics[width=0.25\textwidth]{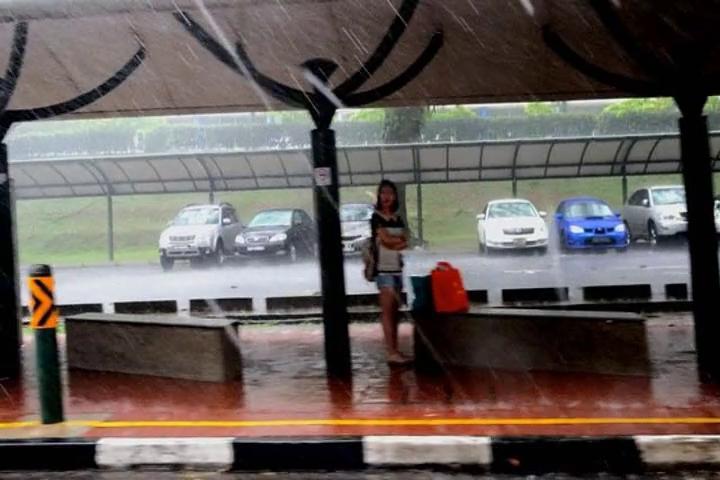} & \includegraphics[width=0.25\textwidth]{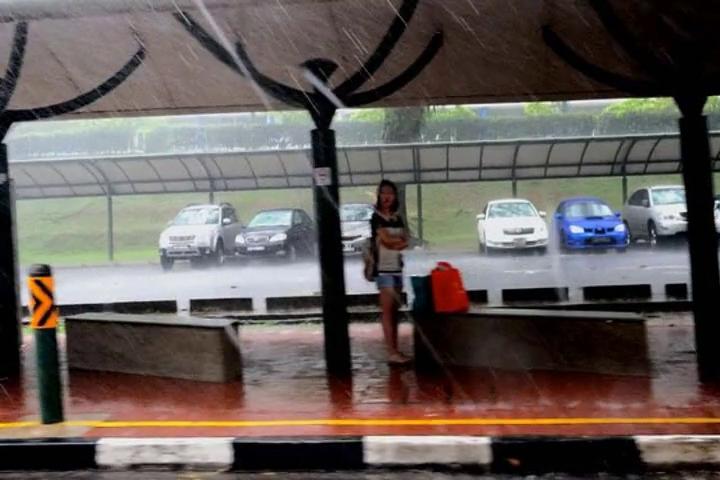}
        \\
        \includegraphics[width=0.25\textwidth]{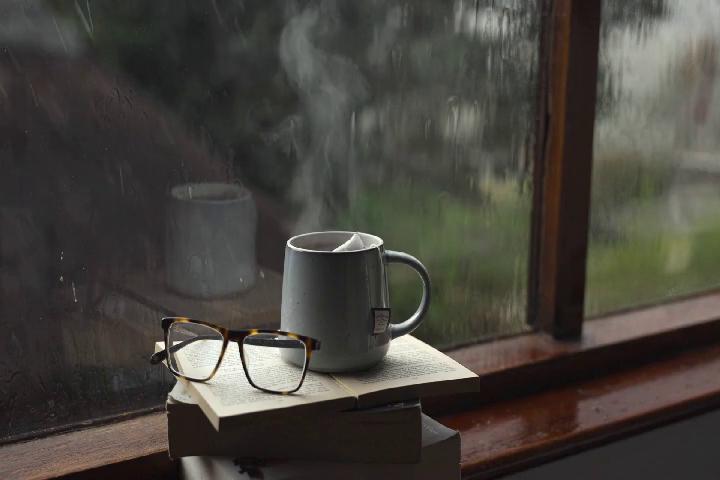} &
        \includegraphics[width=0.25\textwidth]{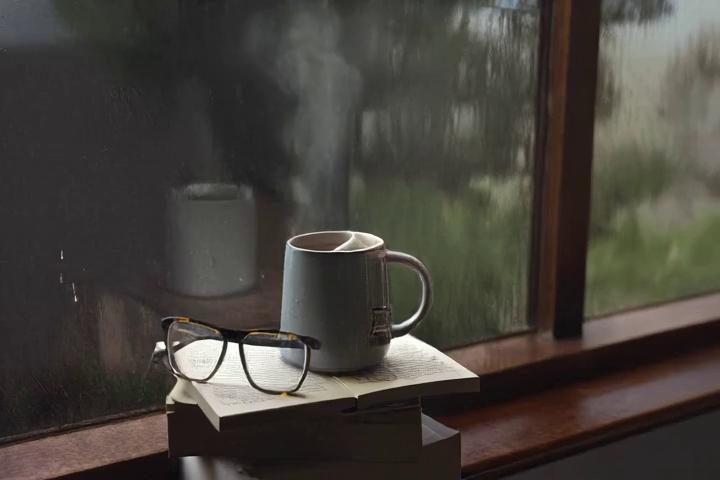} &
        \includegraphics[width=0.25\textwidth]{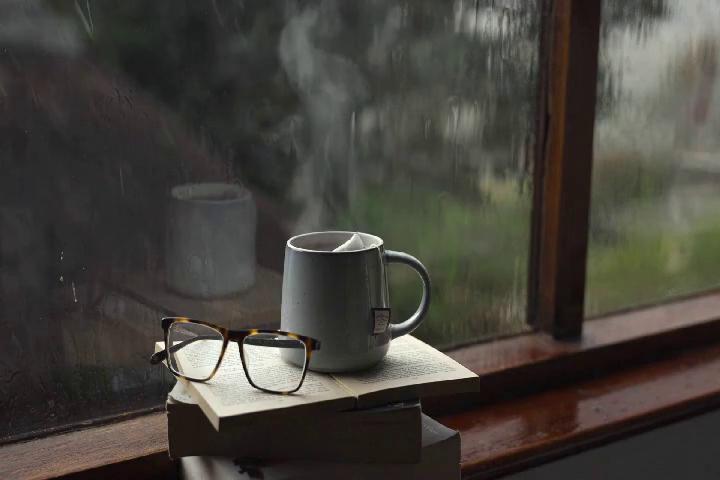} & \includegraphics[width=0.25\textwidth]{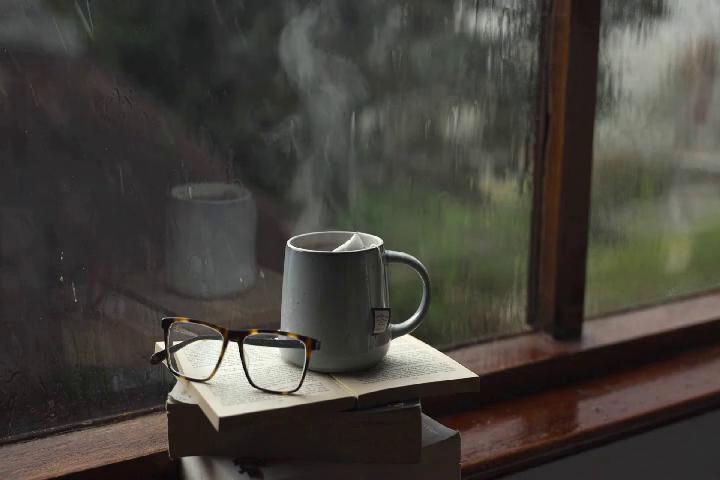}
        \\
        Input & Ours & HistoFormer~\cite{sun2025-histoformer} & T3-DiffWeather~\cite{chen2025-t3_diffweather} \\
    \end{tabular}
\end{subfigure}%
\caption{
Selected frames from derained real-world videos. For best viewing experience see the supplementary video.
}
\label{fig:results3}
\end{figure*}

\section{Results on Synthetic Data}
We test our method on the synthetic test set of NTURain~\cite{chen2018-ntu_rain}, and compare against supervised methods in Table~\ref{tab:ntu_synthetic}. Note that S2VD \cite{yue2021-s2vd}] and RainMamba \cite{wu2024-rainmamba} were trained on NTURain, while ours is training-free, which explains their unfair higher metrics. Histoformer \cite{sun2025-histoformer} and Diff-Plugin \cite{liu2024-diff-plugin} were trained on different synthetic rain datasets and thus generalize less well to NTURain. Our method, despite training-free, performs comparable to other supervised method under cross dataset validation set up. 


\begin{figure*}[t]
    \renewcommand{\arraystretch}{1}
    \small
    \centering
    \setlength{\tabcolsep}{1pt}
    \begin{tabular}{*{5}{>{\centering\arraybackslash}p{0.19\linewidth}}} 
        \multicolumn{5}{c}{
        \includegraphics[width=0.96\linewidth]{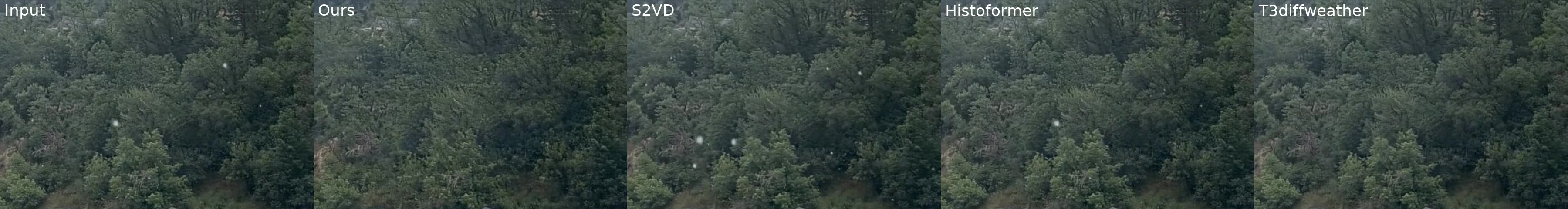}} 
        \\
        \multicolumn{5}{c}{
        \includegraphics[width=0.96\linewidth]{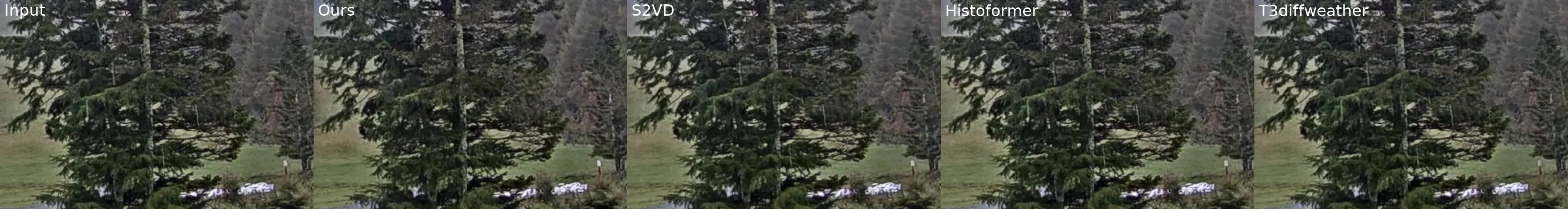}} 
        \\
        Input & Ours & S2VD~\cite{yue2021-s2vd} & Histoformer~\cite{sun2025-histoformer} & T3diffweather~\cite{chen2025-t3_diffweather} 
    \end{tabular}
	\caption{
	\textbf{Qualitative comparison on real rain videos from GT-Rain~\cite{ba2022-gt_rain}.} Please refer to the supplementary for best viewing experience. 
    }
	\label{fig:gtrain}
\end{figure*}

\section{Additional Results on Deraining}
\cref{fig:results3} demonstrates additional results on deraining tasks. Unlike HistoFormer~\cite{sun2025-histoformer} and T3-DiffWeater~\cite{chen2025-t3_diffweather}, the proposed method is able to remove rain from both scenes. \cref{fig:gtrain} shows a qualitative comparison between our method and other baselines on GT-Rain~\cite{ba2022-gt_rain}. The supplementary videos illustrate the difference more clearly. Please open \texttt{\small{videos.html}} to view the video results.

\section{Additional Results on Attention Switching}
\begin{figure*}
    \centering
    \includegraphics[width=0.99\linewidth]{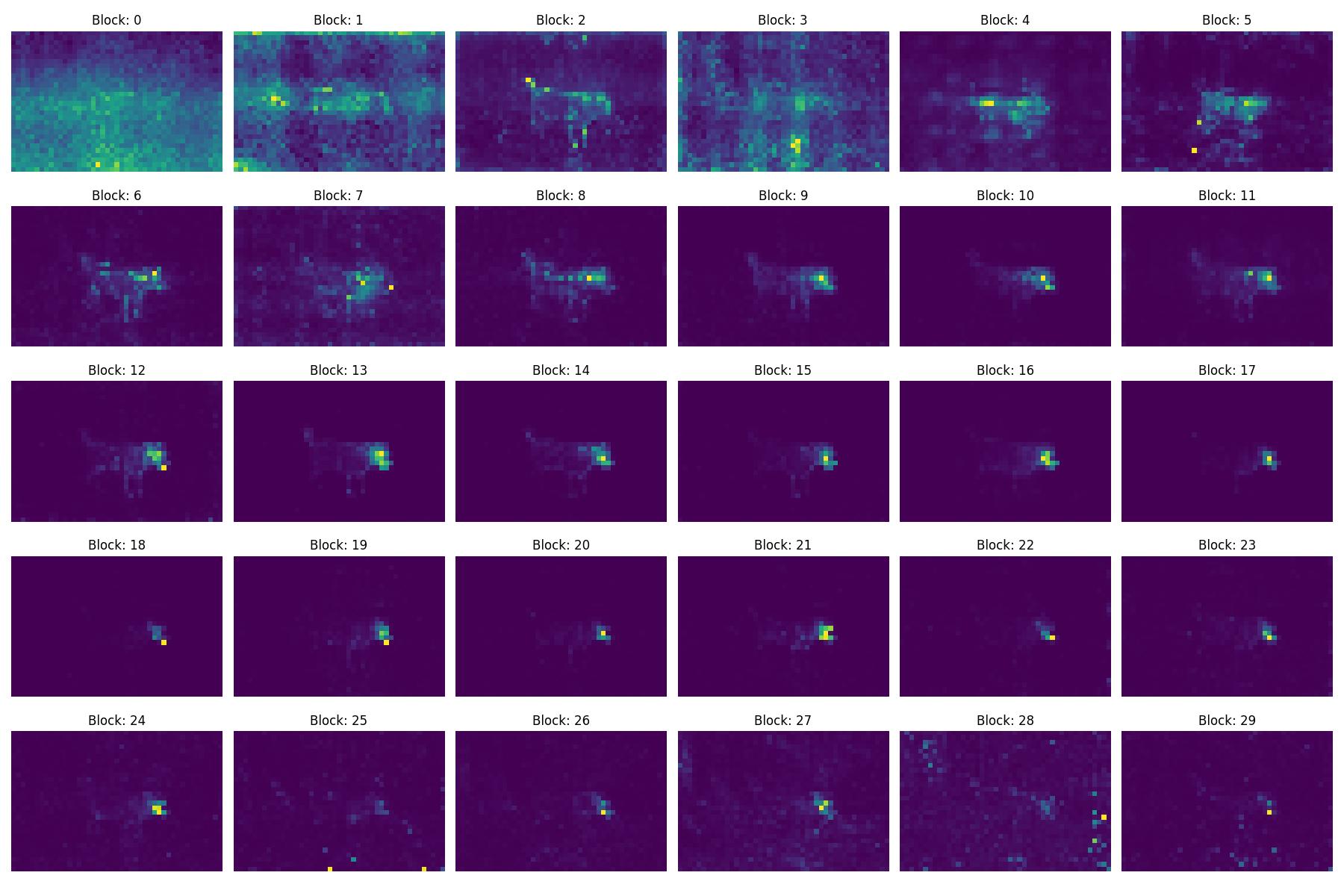}
    \caption{Attention map from the prompt \textit{A dog walking in a rainy forest}. The maps are constructed from the query $Q$ corresponding to the word \textit{dog}, and keys $K$ are from the first latent frame. Note that the first four blocks mainly contain global information, while the last $\sim 15$ blocks contain mostly redundant spatial information.}
    \label{fig:attn_map}
\end{figure*}

\begin{figure*}[t]
\small
\centering
    \setlength{\tabcolsep}{1pt}
        \begin{tabular}{*3c}
            \includegraphics[width=0.32\textwidth]{figures/results/ablation/kv/ra4_Rain/input.jpg} &
            \includegraphics[width=0.32\textwidth]{figures/results/ablation/kv/ra4_Rain/base.jpg} &
            \includegraphics[width=0.32\textwidth]{figures/results/ablation/kv/ra4_Rain/ours.jpg}
            \\
            Input & No blocks used (Base) & Blocks 0-4 and 15-30 (ours) \\
        \end{tabular}
        \begin{tabular}{*3c}
            \rule{0.32\linewidth}{0pt}&
            \includegraphics[width=0.32\textwidth]{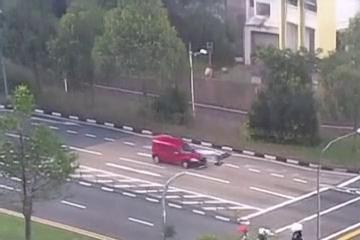} &
            \includegraphics[width=0.32\textwidth]{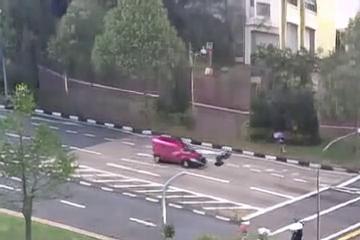}
            \\ 
            & Blocks 15-30 & Blocks 0-4 \\
        \end{tabular}
	\caption{
	Ablation study on different selection of blocks $\mathcal{B}$ for attention switching. Using both the initial four blocks and the later fifteen blocks for attention switching obtains the best results. This can be observed by analyzing the distortions caused by other settings when compared to the input image.
	}
	\label{fig:ablation_blocks2}
\end{figure*}

In Sec.~3.3 of the main paper, we propose using a subset of blocks $\mathcal{B}$ for attention switching to enhance structural preservation. The selection of $\mathcal{B}$ is based on the statistical analysis of high-frequency information in different blocks. We provide an attention map visualization in \cref{fig:attn_map}, showing the attention maps between the prompt \textit{``dog''} and the first frame of the generated latent frame. Note that for the first four blocks, the attention is not localized but focuses more on the global features, while for the last fifteen blocks, the features show redundancy in spatial locality. We perform an ablation study on attention switching across the different blocks in \cref{fig:ablation_blocks2}. When attention switching is not applied in any of the transformer blocks, the result is distorted. In utilizing attention switching in both the initial (the first four blocks) and the last fifteen blocks, the optimal result is obtained.
\cref{fig:ablation_blocks2} (bottom row) shows that using only the initial or the latter blocks results in less optimal results.

\section{Inversion Techniques}
To the best of our knowledge, no previous work has attempted to invert a video using video diffusion models, where the video is represented as a block instead of a set of frames. Frame-based inversion~\cite{cohen2024-slicedit, chen2024-streaming, feng2024-ccedit, geyer2023-tokenflow} methods often lack temporal consistency and hence propose extended attention modules between frames, rely on depth maps and structured noise maps. By using a video diffusion model, the complexity in models can be significantly reduced with regard to temporal consistency.

\begin{figure*}[t]
\small
\centering
\setlength{\tabcolsep}{1pt}
\begin{subfigure}{0.99\textwidth}
\hspace{-0.2cm}
    \begin{tabular}{*4c}
        \raisebox{50pt}[0.04\linewidth][0pt]{\parbox[t]{3mm}{\rotatebox[origin=c]{90}{$t_s = 0$}} } &
        \includegraphics[width=0.32\textwidth]{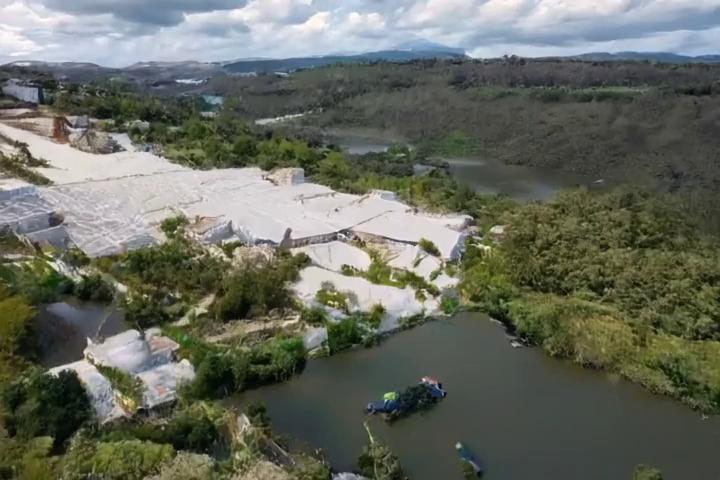} &
        \includegraphics[width=0.32\textwidth]{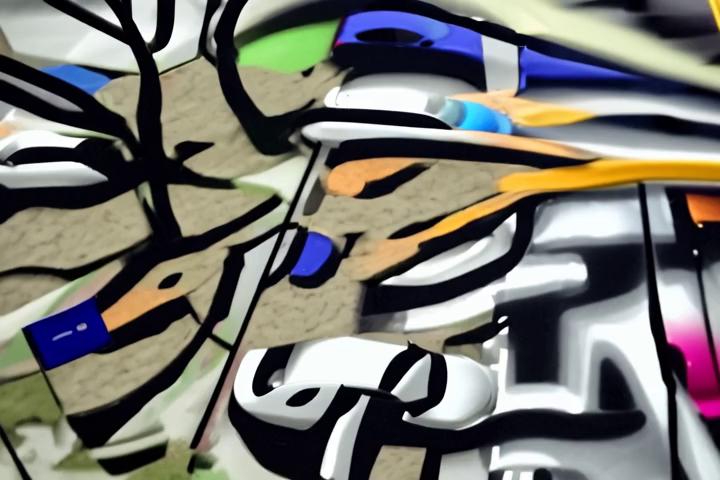} & \includegraphics[width=0.32\textwidth]{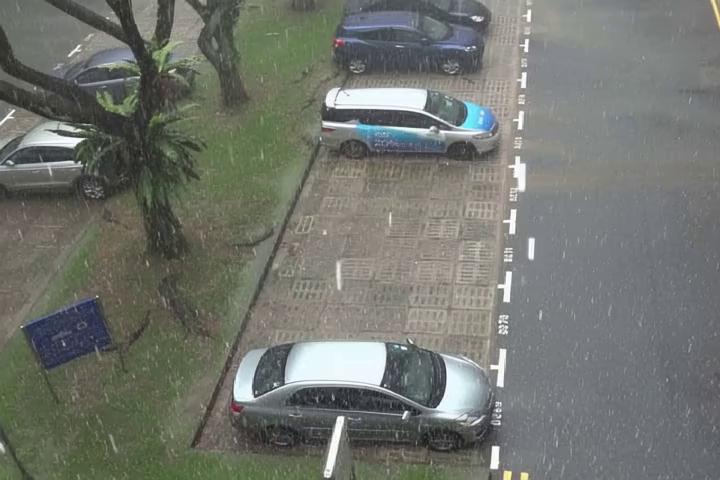}
        \\
         & PSNR=10.4820 & PSNR=10.7400 & PSNR=31.7528
        \\
        \raisebox{50pt}[0.04\linewidth][0pt]{\parbox[t]{3mm}{\rotatebox[origin=c]{90}{$t_s = 25$}} } &
        \includegraphics[width=0.32\textwidth]{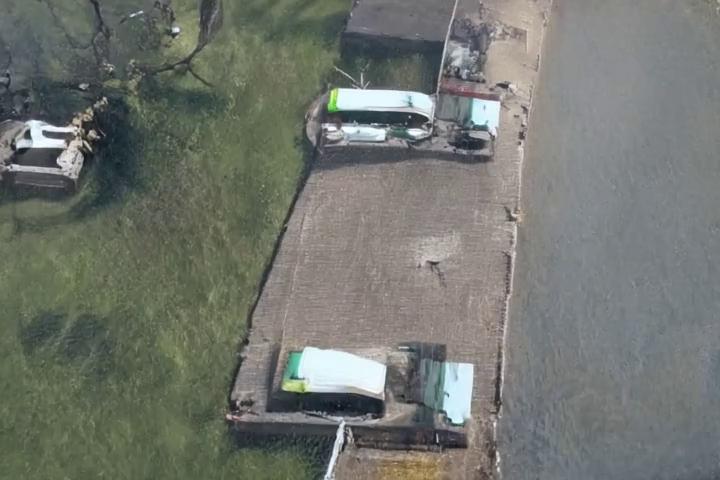} &
        \includegraphics[width=0.32\textwidth]{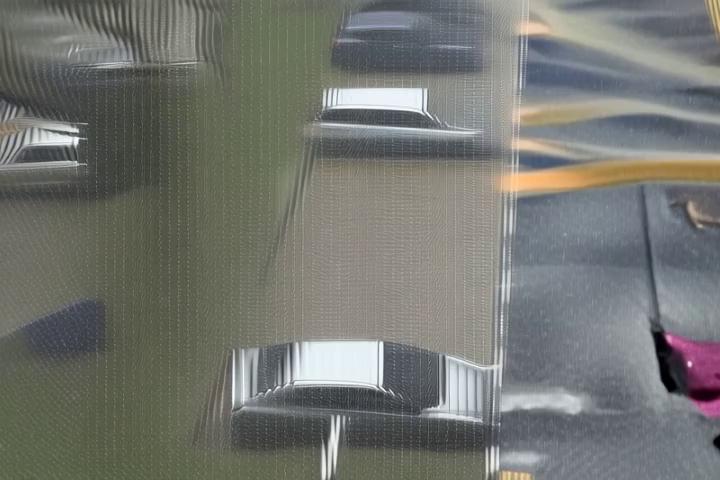} & \includegraphics[width=0.32\textwidth]{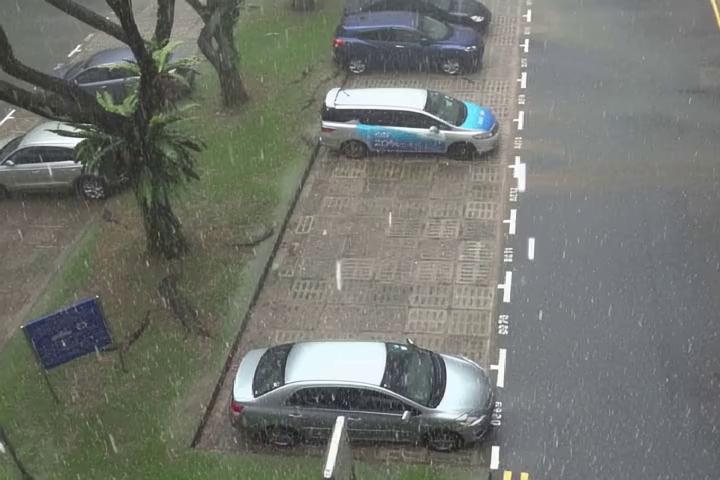}
        \\
         & PSNR=19.5115 & PSNR=20.9285 & PSNR=31.7870
        \\
        \raisebox{50pt}[0.04\linewidth][0pt]{\parbox[t]{3mm}{\rotatebox[origin=c]{90}{$t_s = 40$}} } &
        \includegraphics[width=0.32\textwidth]{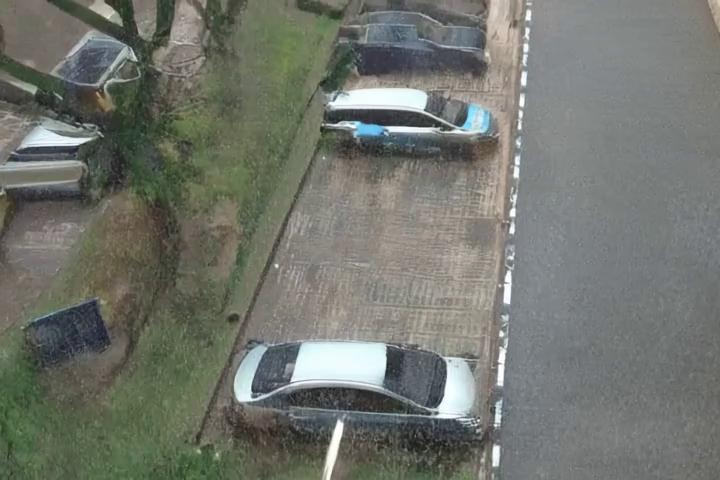} &
        \includegraphics[width=0.32\textwidth]{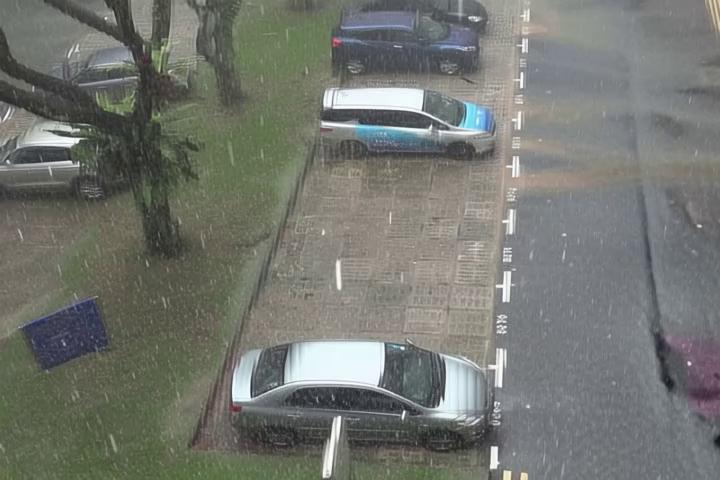} & \includegraphics[width=0.32\textwidth]{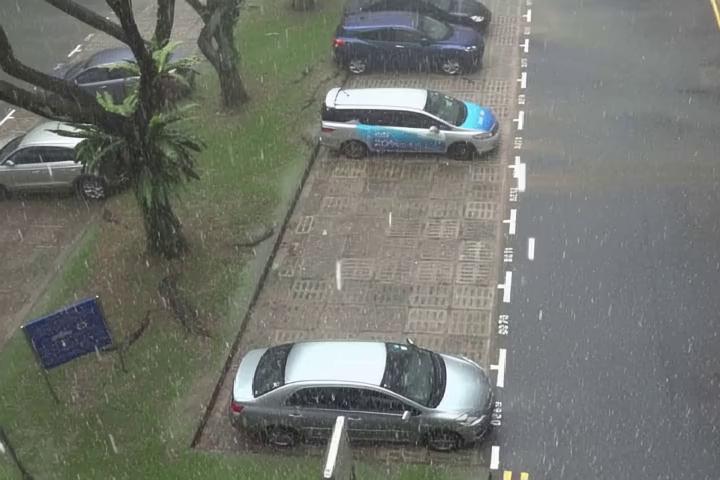}
        \\
         & PSNR=22.7327 & PSNR=27.2412 & PSNR=31.7861
        \\
        \raisebox{50pt}[0.04\linewidth][0pt]{\parbox[t]{3mm}{\rotatebox[origin=c]{90}{$t_s = 50$}} } &
        \includegraphics[width=0.32\textwidth]{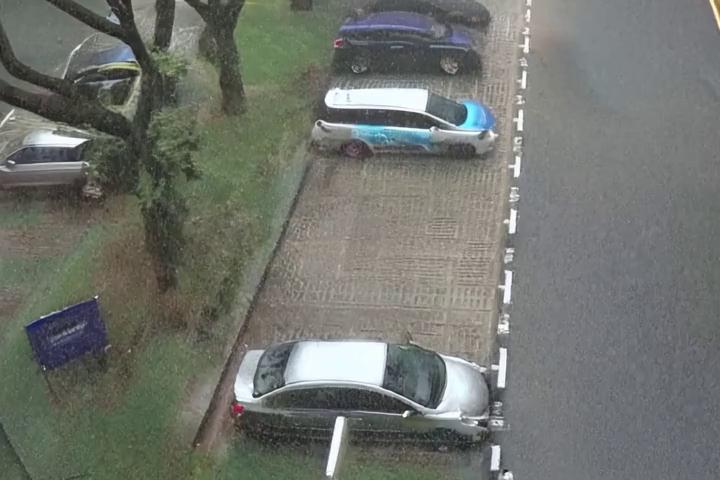} &
        \includegraphics[width=0.32\textwidth]{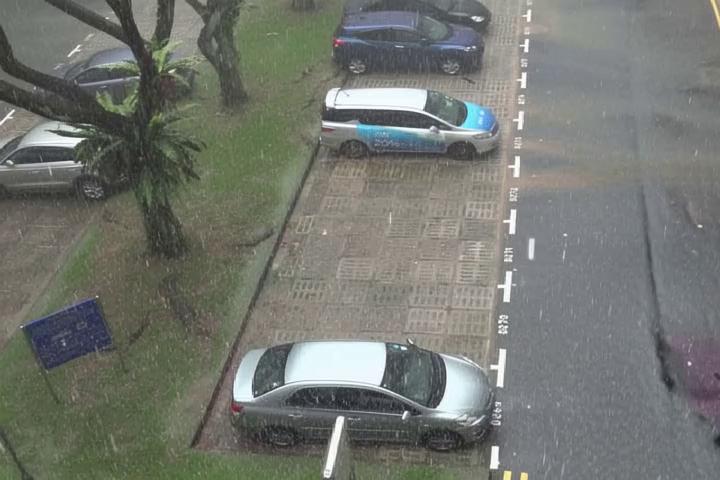} & \includegraphics[width=0.32\textwidth]{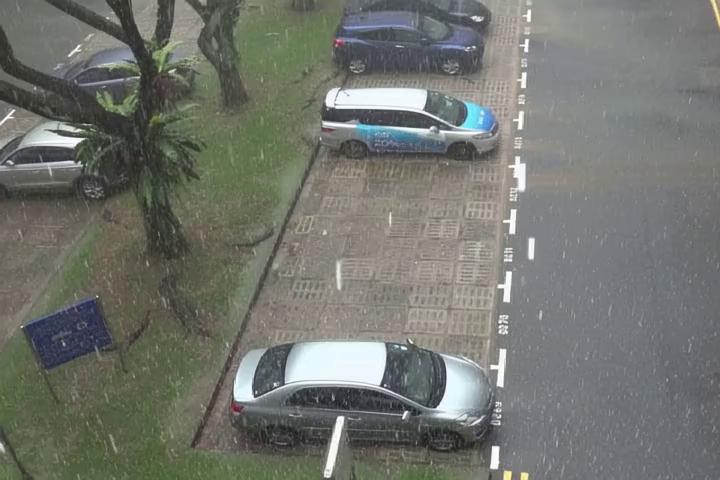}
        \\
         & PSNR=24.3778 & PSNR=30.8184 & PSNR=31.7860
        \\
        \\
         & Video SDEdit inversion & Video DDIM inversion & Video DDPM inversion \\
    \end{tabular}
\end{subfigure}%
\caption{
Comparison of video inversion results at different skip values $t_s$. Video SDEdit inversion loses the entire scene structure at $t_s=0$, while video DDIM inversion can retain some structure from the camera motion and the cars. Video DDPM inversion retains the scene with only a minor loss in high-frequency details. Results improve for higher skip values $t_s$. Note that PSNR is averaged over all video frames.
}
\label{fig:inversion2}
\end{figure*}

We experimented with SDEdit~\cite{meng2021-sdedit}, DDIM inversion~\cite{song2022-ddim}, Null-text inversion~\cite{mokady2023-null_text_inversion} and DDPM inversion~\cite{huberman2024-ddpm_inversion}. As Null-text inversion requires optimization at every time step, the method becomes impractical due to a long runtime of 50 minutes for a single video. Hence, we have not included it in this comparison. The results are shown in \cref{fig:inversion2}. Both video SDEdit and video DDIM inversion struggle to retain details for full inversion. An inversion that starts from $t_s$ is more practical since these values can be used for modifying the content. At $t_s = 25$, both approaches can get the global scene structure but are still unable to produce scene details. Video DDPM inversion is capable of reconstructing the scene with fine-grained details even from the initial timestep.

\section{Additional Results on Desnowing}
\begin{figure*}[t]
\small
\centering
\setlength{\tabcolsep}{1pt}
\begin{subfigure}{0.99\textwidth}
\hspace{-0.2cm}
    \begin{tabular}{*4c}
        \includegraphics[width=0.25\textwidth]{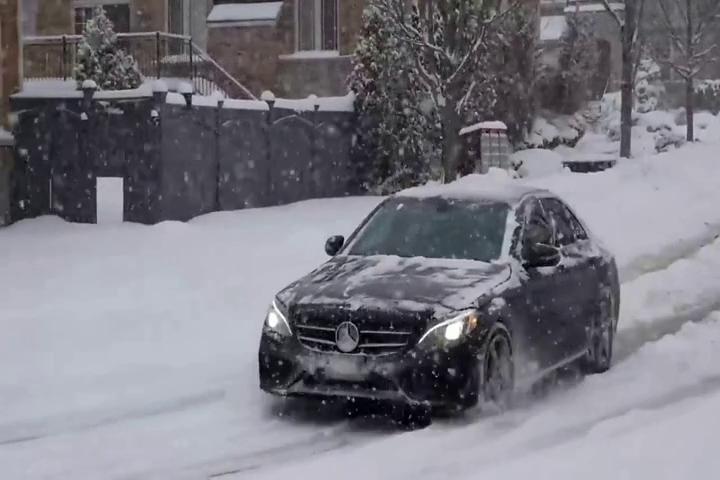} &
        \includegraphics[width=0.25\textwidth]{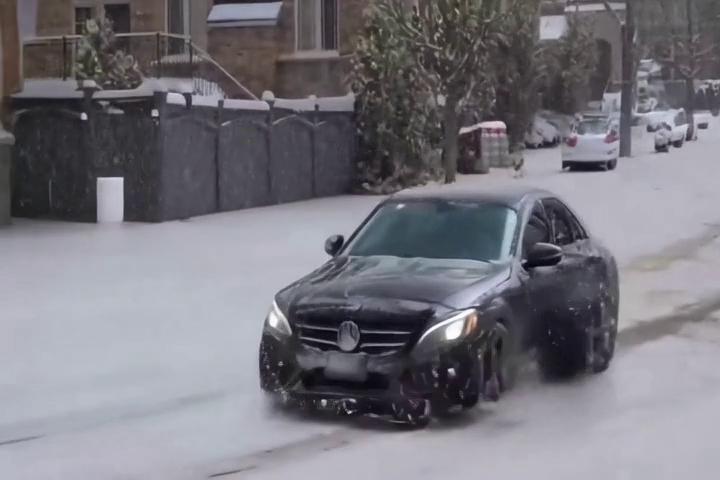} &
        \includegraphics[width=0.25\textwidth]{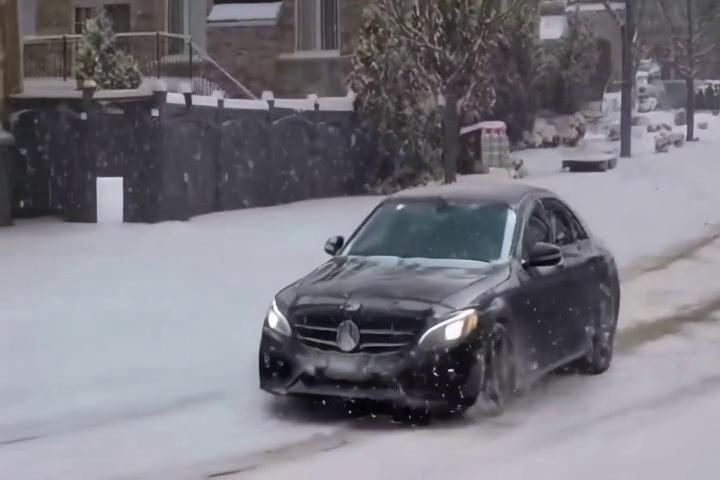} & \includegraphics[width=0.25\textwidth]{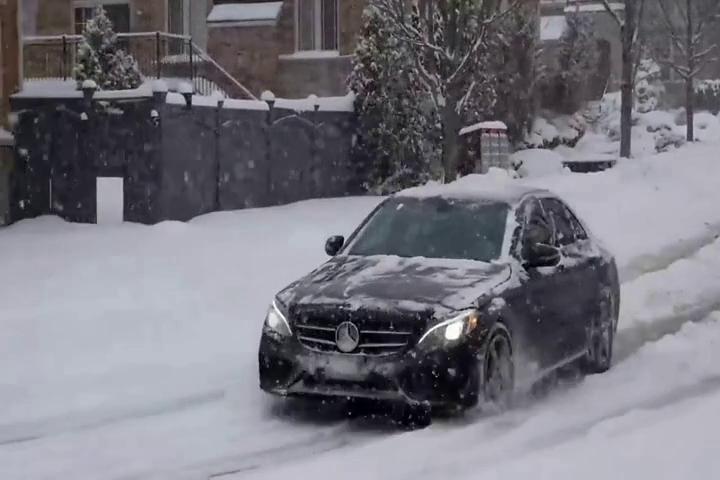}
        \\
        \includegraphics[width=0.25\textwidth]{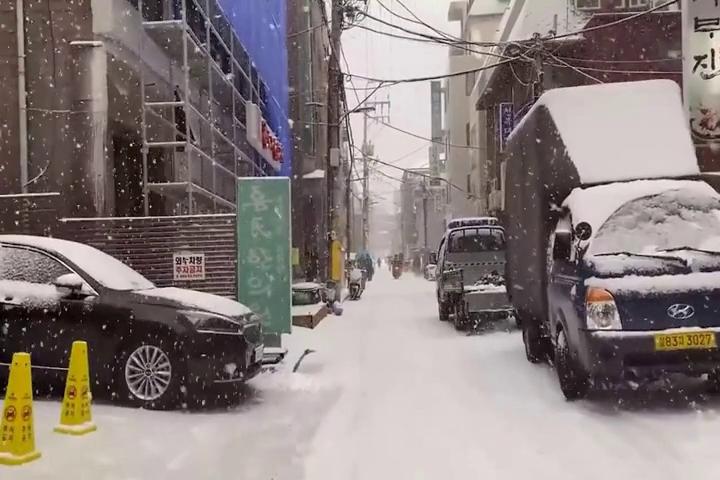} &
        \includegraphics[width=0.25\textwidth]{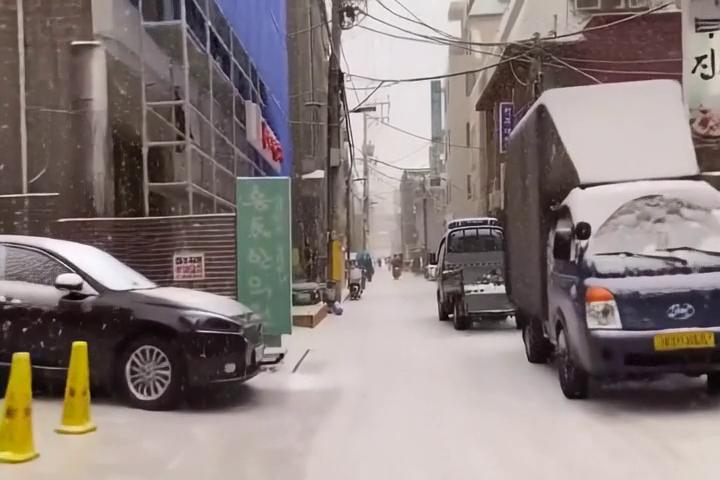} &
        \includegraphics[width=0.25\textwidth]{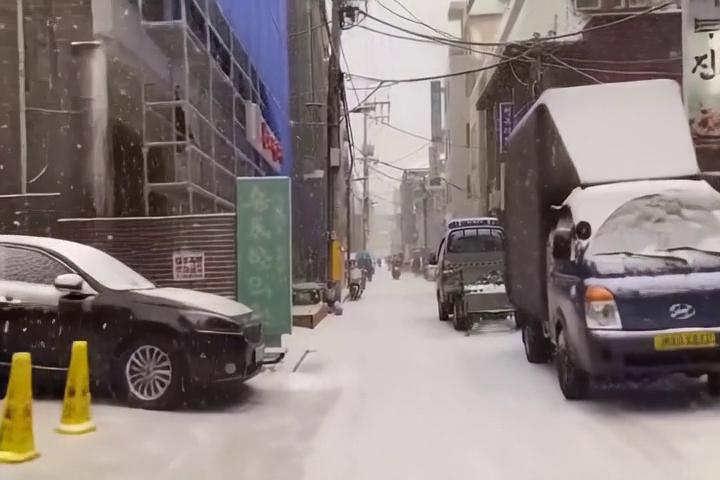} & \includegraphics[width=0.25\textwidth]{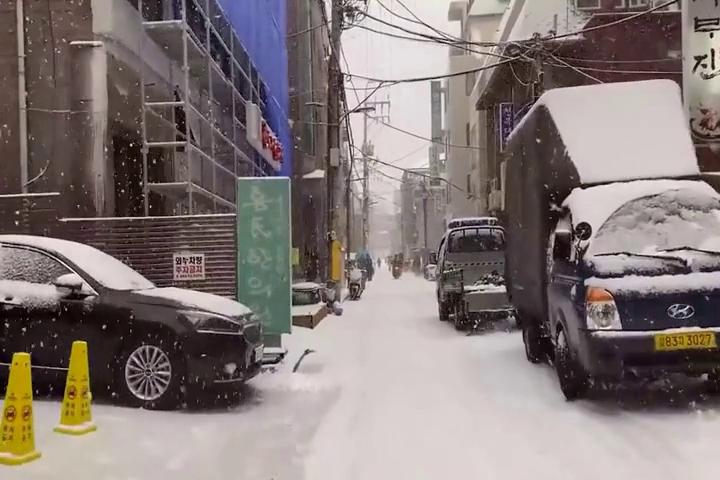}
        \\
        \includegraphics[width=0.25\textwidth]{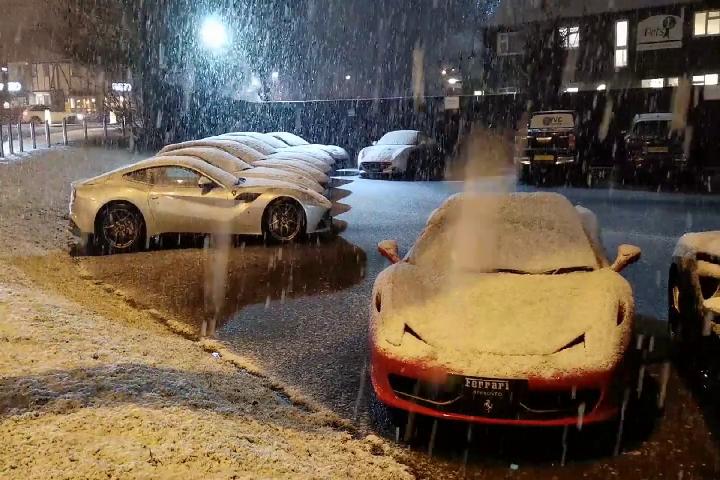} &
        \includegraphics[width=0.25\textwidth]{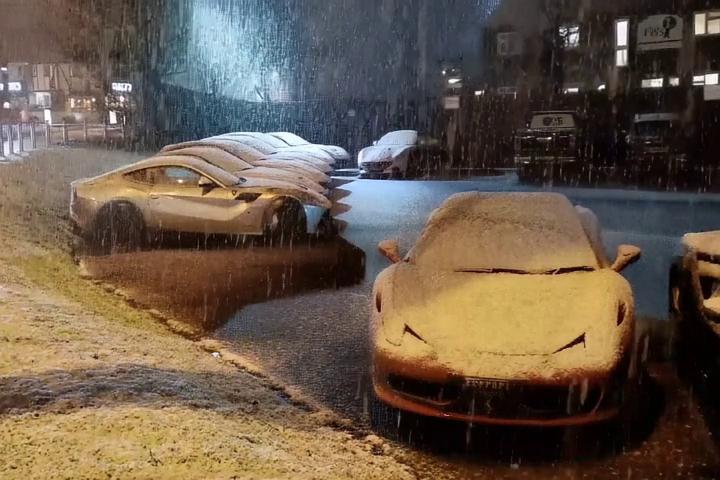} &
        \includegraphics[width=0.25\textwidth]{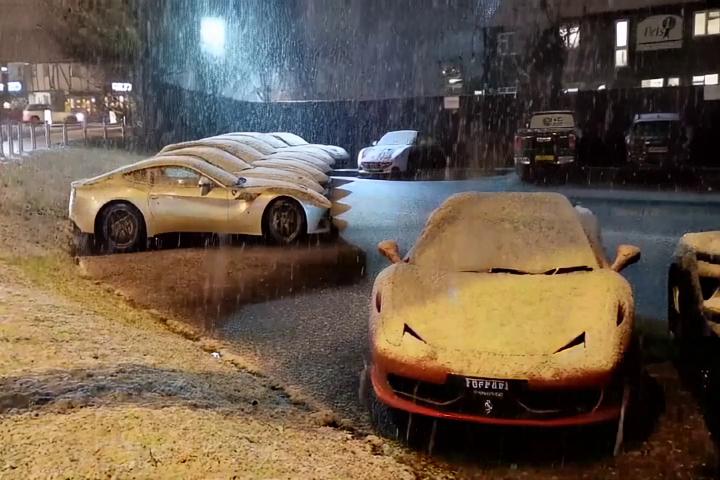} & \includegraphics[width=0.25\textwidth]{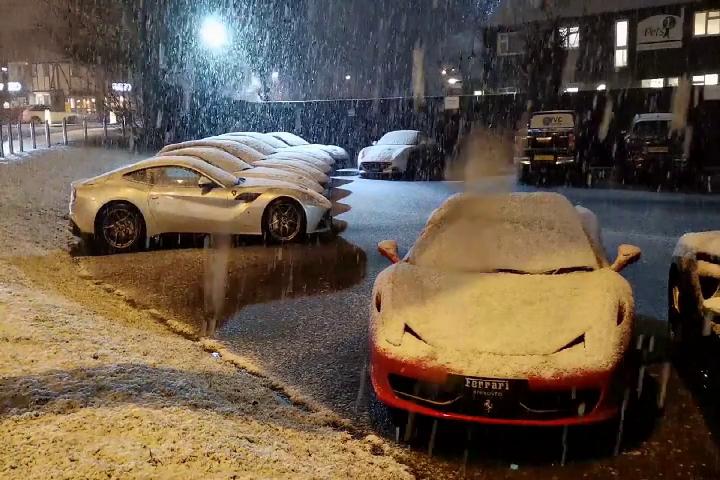}
        \\
        \includegraphics[width=0.25\textwidth]{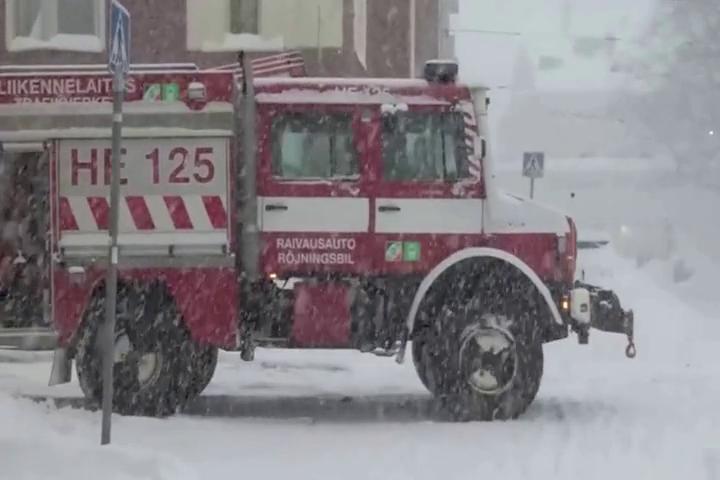} &
        \includegraphics[width=0.25\textwidth]{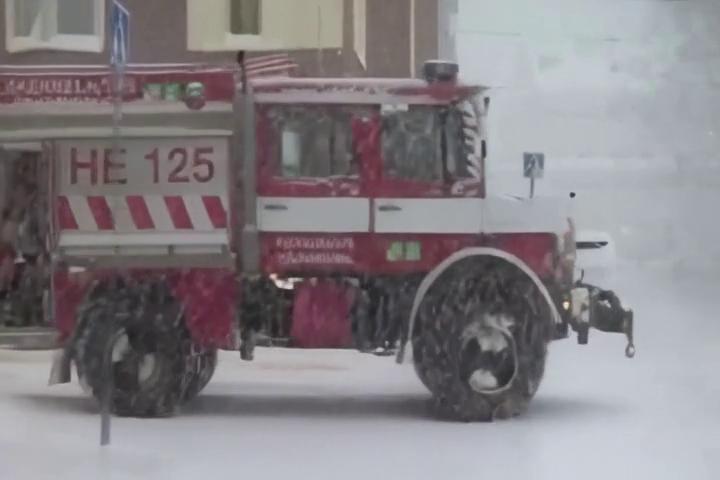} &
        \includegraphics[width=0.25\textwidth]{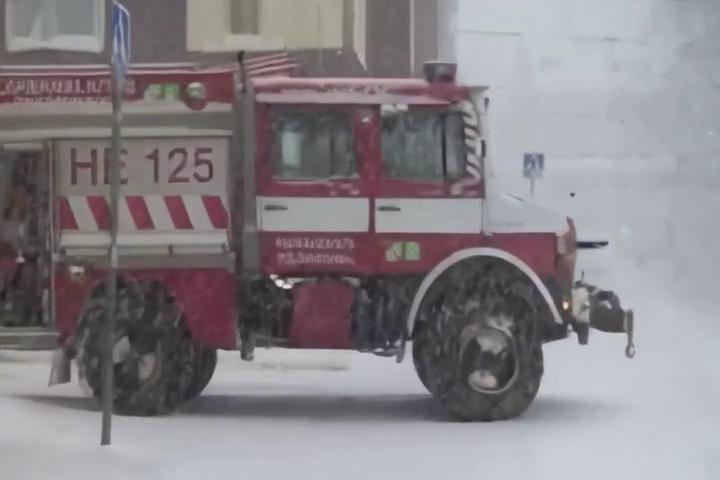} & \includegraphics[width=0.25\textwidth]{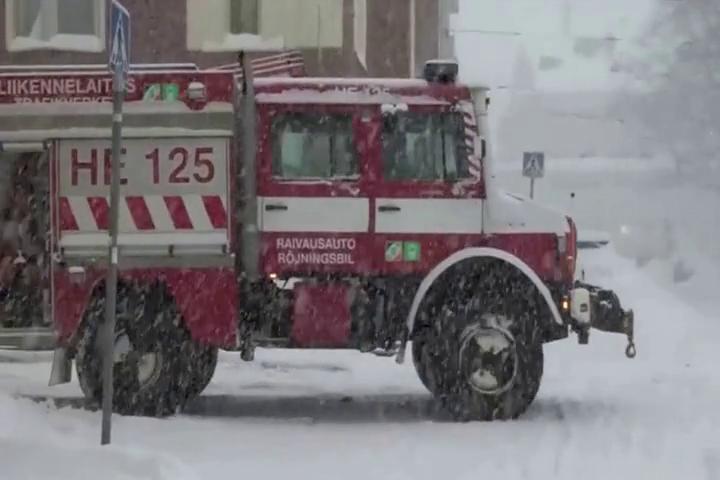}
        \\
        Input & Ours & Base (Ours) & TURTLE~\cite{ghasemabadi2024-turtle} \\
    \end{tabular}
\end{subfigure}%
\caption{
Selected frames from desnowed real-world videos. Base refers to the case without attention switching. For the best viewing experience, see the supplementary material.
}
\label{fig:results_snow}
\end{figure*}

Desnowing~\cite{chen2023-rvsd} is the task of removing snow, akin to deraining. The problem has been less studied, likely due to less available data and posing issues less frequently for applications. We experimented with our approach on snow in \cref{fig:results_snow} with samples collected from the internet and RealSnow85~\cite{wu2025-realsnow85}. The proposed method is able to better remove snow compared to the state-of-the-art method TURTLE~\cite{ghasemabadi2024-turtle}. However, compared to rainy cases, the proposed method is less effective at removing all of the snow and sometimes struggles with structural preservation. 

\begin{figure*}[t]
\small
\centering
\setlength{\tabcolsep}{1pt}
\begin{subfigure}{0.99\textwidth}
\hspace{-0.2cm}
    \begin{tabular}{*3c}
        \includegraphics[width=0.33\textwidth]{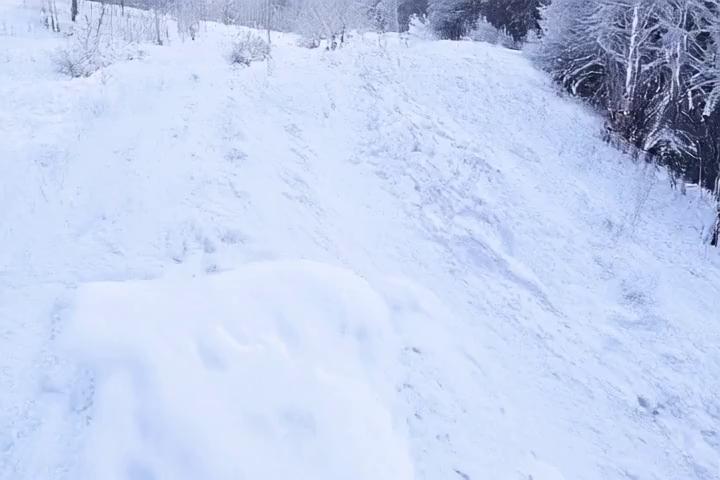} &
        \includegraphics[width=0.33\textwidth]{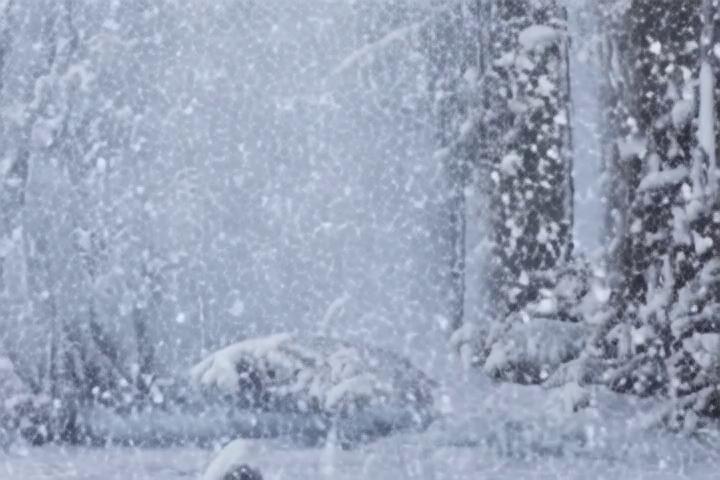} &
        \includegraphics[width=0.33\textwidth]{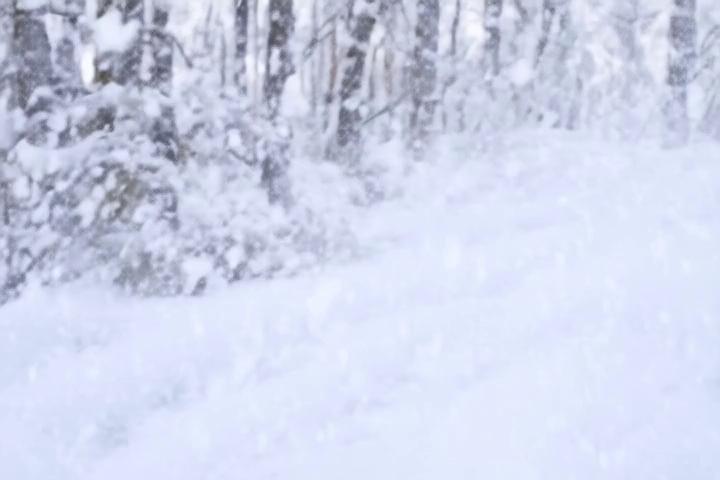}
        \\
        \textit{``snow''} & \textit{``snowing''} & \textit{``light snowing''} \\
    \end{tabular}
\end{subfigure}%
\caption{
Visualization of different snow prompts. \textit{Left:} The result generated with the prompt \textit{``snow''} produces snow on the ground instead of a falling snow effect. \textit{Middle:} In using the prompt \textit{``snowing''}, the model generates falling snow. \textit{Right:} Unlike the results in \cref{fig:rain_prompt} for the prompt \textit{light}, the same prompt affects snow generation differently. Note that the generated prompt is entangled with a snowy forest.
}
\label{fig:snow_prompts}
\end{figure*}

After performing a similar analysis to that in \cref{fig:rain_prompt}, we find that the base model CogVideoX has not properly disentangled the concept of snow. \cref{fig:snow_prompts} shows how the snow prompt generates a forest background, whereas the rain prompt in Fig.~5 generates no background. We hypothesize that the forest background is due to the training data, where snowy scenes mainly contain a forest in the background. This property harms the desnowing process, as the score estimate $\hat{\epsilon}_\theta(x_t)$ is not only pushed away from the snowy concept but also the forest concept, leading to worse structure preservation. Using larger video diffusion models in the future would likely disentangle the concepts better, potentially improving different restoration tasks, e.g., desnowing.

{
    \small
    \bibliographystyle{ieeenat_fullname}
    \bibliography{main}
}
\end{document}